\documentclass[sigconf]{acmart}
\usepackage{microtype}
\usepackage{graphicx}
\usepackage{subfigure}
\usepackage{booktabs} 
\usepackage{multirow}
\usepackage{amsmath}
\usepackage{xcolor}
\usepackage{marvosym}

\usepackage{hyperref}

\usepackage{url}

\usepackage{amsmath}
\usepackage{mathtools}
\usepackage{amsthm}
\usepackage[capitalize,noabbrev]{cleveref}
\usepackage{lipsum}

\theoremstyle{plain}

\theoremstyle{definition}

\theoremstyle{remark}

\usepackage[textsize=tiny]{todonotes}

\begin{document}
\title{Understanding Privacy Risks of Embeddings Induced by Large Language Models}
\author{Zhihao Zhu, Ninglu Shao, Defu Lian, Chenwang Wu, Zheng Liu, Yi Yang, Enhong Chen}
\begin{abstract}
Large language models (LLMs) show early signs of artificial general intelligence but struggle with hallucinations. One promising solution to mitigate these hallucinations is to store external knowledge as embeddings, aiding LLMs in retrieval-augmented generation. However, such a solution risks compromising privacy, as recent studies experimentally showed that the original text can be partially reconstructed from text embeddings by pre-trained language models. The significant advantage of LLMs over traditional pre-trained models may exacerbate these concerns. To this end, we investigate the effectiveness of reconstructing original knowledge and predicting entity attributes from these embeddings when LLMs are employed. Empirical findings indicate that LLMs significantly improve the accuracy of two evaluated tasks over those from pre-trained models, regardless of whether the texts are in-distribution or out-of-distribution. This underscores a heightened potential for LLMs to jeopardize user privacy, highlighting the negative consequences of their widespread use. We further discuss preliminary strategies to mitigate this risk.

\end{abstract}

\maketitle

\section{Introduction}

Large language models~\cite{chang2023survey, kaddour2023challenges} have garnered significant attention for their exceptional capabilities across a wide range of tasks like natural language generation~\cite{axelsson2023using, min2023recent}, question answering~\cite{zhu2021retrieving, lu2022learn}, and sentiment analysis~\cite{albadani2022novel, wankhade2022survey}. 

Nonetheless, it's observed that large language models can confidently assert non-existent facts during their reasoning process. For example, Bard, Google's AI chatbot, concocted information in the first demo that the James Webb Space Telescope had taken the first pictures of a planet beyond our solar system~\cite{coulter2023alphabet}. Such a hallucination problem~\cite{zhang2023siren, li2023halueval} of large language models is a significant barrier to artificial general intelligence~\cite{goertzel2014artificial, pei2019towards}. A primary strategy for tackling the issue of hallucinations is to embed external knowledge in the form of embeddings into a vector database~\cite{guo2022manu, friedman2020symbolic}, making them accessible for retrieval augmented generation by large language models~\cite{cui2023chatlaw, andriopoulos2023augmenting}.

An embedding model~\cite{patil2023survey, selva2021review} encodes the original objects' broad semantic information by transforming the raw objects (e.g., text, image, user profile) into real-valued vectors of hundreds of dimensions. The advancement of large language models enhances their ability to capture and represent complex semantics more effectively, such that an increasing number of businesses (e.g., OpenAI~\cite{neelakantan2022text} and Cohere~\cite{cohere}) have launched their embedding APIs based on large language models. 
Since embeddings are simply real-valued vectors, it is widely believed that it is challenging to decipher the semantic information they contain. Consequently, embeddings are often viewed as secure and private, as noted by ~\cite{song2020information}, leading data owners to be less concerned about safeguarding the privacy of embeddings compared to raw external knowledge. However, in recent years, multiple studies~\cite{song2020information, li2023sentence, morris2023text} have highlighted the risk of embeddings compromising privacy. More specifically, the pre-trained LSTM (Long Short-Term Memory) networks~\cite{hochreiter1997long} or other language models can recover parts of the original texts and author information from text embeddings, which are generated by open-source embedding models. Although current studies have exposed security weaknesses in embeddings, the effects of large language models on the privacy of these embeddings have not been fully explored. A pressing issue is whether LLMs' emergent capabilities enable attackers to more effectively decipher sensitive information from text embeddings. This issue is driven not only by the proliferation of large language models but also by the availability of the embedding APIs based on LLMs, which permits attackers to gather numerous text-embedding pairs to build their attack models.

To this end, we establish a comprehensive framework that leverages a large language model (LLM) to gauge the potential privacy leakage from text embeddings produced by the open-sourced embedding model. From a security and privacy perspective, LLM serves as the attacker, and the embedding model acts as the target, while the goal is to employ the attack model to retrieve sensitive and confidential information from the target model. Specifically, our approach begins with fine-tuning attack models to enable text reconstruction from the outputs of the target model. Following this, we assess the privacy risks of embeddings via two types of attack scenarios. On the one hand, we recover the texts from their embeddings in both in-distribution and out-of-distribution scenarios. On the other hand, we identify certain private attributes of various entities in the original text (such as birthdays, nationalities, criminal charges, etc.) and predict these attributes from the text embeddings. This prediction is determined by the attribute that exhibits the highest cosine similarity between the text embedding and the corresponding attribute embedding. Consequently, this method does not necessitate training with supervised data. Should the target embedding model decline to generate embeddings for attributes with extremely brief texts described (1-2 words) out of embedding stealing concerns, we introduce an external embedding model that acts as a proxy to project the original text and the attribute value in the same embedding space. Specifically, this external model is tasked with embedding the attribute and the text reconstructed by the attack model, the latter being derived from text embeddings produced by the target embedding model.

The evaluation of text reconstruction reveals that 1) a larger attack language model, when fine-tuned with a sufficient amount of training data, is capable of more accurately reconstructing texts from their embeddings in terms of metrics like BLEU~\cite{papineni2002bleu}, regardless of whether the texts are in-distribution or out-of-distribution; 2) in-distributed texts are more readily reconstructed than out-of-distributed texts, with the reconstruction accuracy for in-distributed texts improving as the attack model undergoes training with more data; 3) the attack model can improve the reconstruction accuracy as the expressiveness of the target embedding models increases.

The evaluation of attribute prediction demonstrates that 1) attributes can be predicted with high accuracy across various domains, including encyclopedias, news, medical, and legislation. This means the attacker is capable of inferring details like a patient's health condition, a suspect's criminal charges, and an individual's birthday from a set of seemingly irrelevant digital vectors, highlighting a significant risk of privacy leakage; 2) generally speaking, enlarging the scale of the external/target embedding model substantially enhances the accuracy of attribute prediction; 3) when the target model denies embedding services for very short texts, the most effective approach using reconstructed text by the attack model can achieve comparable performance to using original text, when the target model and the external embedding model are configured to be the same.

From the experiments conducted, we find that knowledge representations merely through numerical vectors encompass abundant semantic information. The powerful generative capability of large language models can continuously decode this rich semantic information into natural language. If these numerical vectors contain sensitive private information, large language models are also capable of extracting such information. The development trend of large language models is set to increase these adverse effects, underscoring the need for our vigilance. Our research establishes a foundation for future studies focused on protecting the privacy of embeddings. For instance, the finding that accuracy in text reconstruction diminishes with increasing text length indicates that lengthening texts may offer a degree of privacy protection. Furthermore, the ability of the attack model to reconstruct out-of-distributed texts points towards halting the release of original texts associated with released embeddings as a precaution.

\section{Main Results}
\subsection{Fine-Tuning the Attack Model}
\begin{figure}[htbp]
    \centering
    \includegraphics[width=\columnwidth]{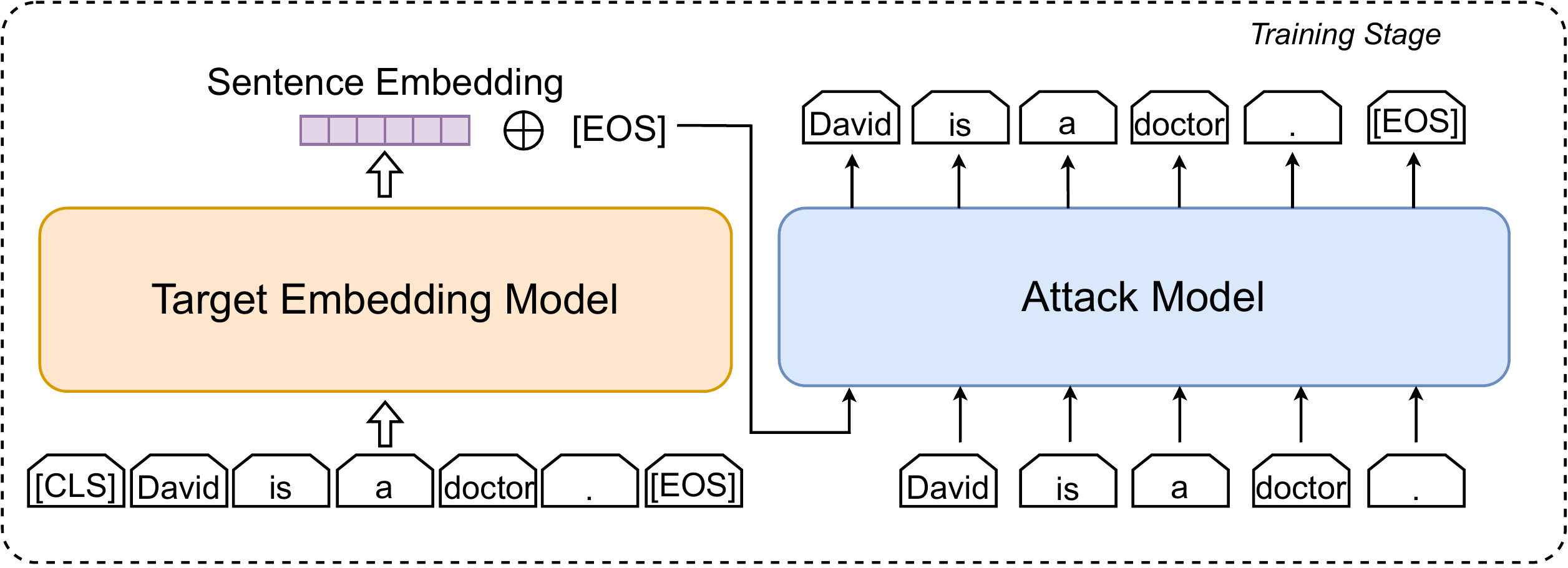}
    \vspace{-1em}
    \caption{\textbf{The fine-tuning of the foundation attack model.}
    Initially, the attacker queries the target embedding model to convert the collected text into text embeddings. To signify the completion of the embedding input, an EOS (End-of-Sentence) token is appended to the text embedding. Next, the attacker selects the pre-trained GPT2 model as the attack model and uses the collected text and corresponding text embeddings as a dataset to train the attack model. When a text embedding is input, the attack model is trained to sequentially reconstruct the related original text.}
    \label{workflow1}
\end{figure}

\begin{table}[h]
\centering
\small
\caption{Sizes of attack models and embedding models}\label{model}
\begin{tabular}{l|ccc}
\toprule
Attack Model     &  GPT2 & GPT2-Large & GPT2-XL\\
\midrule
\#parameters & 355M & 744M & 1.5B\\
\toprule
Target Model     &  SimCSE & BGE-Large-en & E5-Large-v2\\
\midrule
\#parameters & 110M & 326M & 335M\\
\bottomrule
\end{tabular}
\end{table}
We employ pre-trained GPT2~\cite{radford2019language} of varying sizes as the attacking model to decipher private information from embeddings produced by the target embedding models such as SimCSE~\cite{gao2021simcse}, BGE-Large-en~\cite{bge_embedding}, and E5-Large-v2~\cite{wang2022text}. We hypothesize that larger embedding models, due to their capacity to capture more information, are more likely to be exposed to a greater privacy risk. Consequently, all these models are designated as the target models, and their numbers of parameters are detailed in Table~\ref{model}. It's important to note that we treat the target model as a black box, meaning we do not have access to or knowledge of its network architecture and parameters. Figure~\ref{workflow1} showcases the fine-tuning process for the attack model. Initially, the example text ``David is a doctor.'' is inputted into the target embedding model to generate its respective embedding. This embedding is then used as the input for the attack model, which aims to reconstruct the original text based solely on this embedding. An EOS (End-of-Sentence) token is appended to the text embedding to signal the end of the embedding input. The attacker's training goal is to predict the t-th token of the original text based on text embedding and the preceding (t-1) tokens of the original text. In the testing phase, the attacker employs beam search~\cite{freitag2017beam} to progressively generate tokens up to the occurrence of the EOS. Once the attack model has been fine-tuned, we evaluate the privacy risks of embeddings through two distinct attack scenarios: text reconstruction and attribute prediction.

\subsection{Evaluation of Text Reconstruction}
\begin{table*}[t]
\centering
\caption{Reconstruction attack performance against different embedding models on the wiki dataset. The best results are highlighted in bold. $*$ represents that the advantage of the best-performed attack model over other models is statistically significant (\textit{p}-value $<$ 0.05).}
\scalebox{0.79}{
\begin{tabular}{ll|cc|cc|cc}
\hline
\multicolumn{2}{l|}{Training data}              & \multicolumn{2}{c|}{wiki-small} & \multicolumn{2}{c|}{Wiki-large} & \multicolumn{2}{c}{Wiki-xl} \\
\hline
Target model               & Attack   model       & BLEU-1                  & ROUGE-1                 & BLEU-1                  & ROUGE-1                 & BLEU-1                  & ROUGE-1                 \\
\hline
\multirow{3}{*}{SimCSE}       & GPT2        & 0.3184±0.0010          & \textbf{0.3212±0.0010}$^*$ & \textbf{0.4512±0.0010}$^*$ & \textbf{0.4961±0.0011}$^*$ & 0.4699±0.0016          & 0.5256±0.0010          \\
                              & GPT2\_large & 0.2996±0.0014          & 0.2913±0.0012          & 0.4349±0.0013          & 0.4678±0.0009          & 0.5293±0.0010          & 0.5930±0.0009          \\
                              & GPT2\_xl    & \textbf{0.3196±0.0011}$^*$ & 0.3112±0.0013          & 0.4455±0.0015          & 0.4833±0.0010          & \textbf{0.5331±0.0011}$^*$ & \textbf{0.5987±0.0007}$^*$ \\
\hline
\multirow{3}{*}{BGE-Large-en} & GPT2        & \textbf{0.3327±0.0014}$^*$ & \textbf{0.3288±0.0011}$^*$ & 0.4173±0.0016          & 0.4483±0.0009          & 0.4853±0.0011          & 0.5337±0.0016          \\
                              & GPT2\_large & 0.2935±0.0013          & 0.2783±0.0011          & 0.4446±0.0011          & 0.4788±0.0006          & 0.5425±0.0012          & 0.5998±0.0010          \\
                              & GPT2\_xl    & 0.3058±0.0019          & 0.3043±0.0012          & \textbf{0.4689±0.0011}$^*$ & \textbf{0.5057±0.0007}$^*$ & \textbf{0.5572±0.0008}$^*$ & \textbf{0.6151±0.0007}$^*$ \\
\hline
\multirow{3}{*}{E5-Large-v2}  & GPT2        & \textbf{0.3329±0.0013}$^*$ & \textbf{0.3341±0.0012}$^*$ & 0.4838±0.0005          & 0.5210±0.0008          & 0.5068±0.0016          & 0.5522±0.0014          \\
                              & GPT2\_large & 0.3093±0.0009          & 0.2875±0.0012          & 0.4700±0.0011          & 0.4990±0.0010          & 0.5679±0.0011          & 0.6220±0.0011          \\
                              & GPT2\_xl    & 0.3083±0.0013          & 0.3017±0.0013          & \textbf{0.4993±0.0013}$^*$ & \textbf{0.5274±0.0011}$^*$ & \textbf{0.5787±0.0007}$^*$ & \textbf{0.6378±0.0009}$^*$ \\
\hline
\end{tabular}}
\label{main}
\end{table*}

\subsection*{Settings}
For each text in the test set, we reconstruct it using the attack model based on its embedding generated by the target model. To evaluate the reconstruction accuracy, we employ two metrics: BLEU (Bilingual Evaluation Understudy)~\cite{papineni2002bleu} and ROUGE (Recall-Oriented Understudy for Gisting Evaluation)~\cite{lin2004rouge}. 
Specifically, we utilize BLEU-1 and ROUGE-1, which are based solely on unigrams, as they yield better results compared to BLEU and ROUGE based on other n-grams~\cite{al2017arabic}. These metrics gauge the similarity between the original and reconstructed texts. Given that the temperature setting influences the variability of the text produced by GPT, a non-zero temperature allows for varied reconstructed texts given the same text embedding. We calculate the reconstruction accuracy across 10 trials to obtain mean and standard error for statistical analysis. Based on these outcomes, we compare the performance of various attack and target configurations using a two-sided unpaired t-test~\cite{duncan1975t}. The evaluation is conducted on seven datasets, including wiki~\cite{saez2020wikimedia}, wiki-bio~\cite{DBLP:journals/corr/LebretGA16}, cc-news~\cite{Hamborg2017}, pile-pubmed~\cite{gao2020pile}, triage~\cite{li2021topic}, cjeu-terms~\cite{chalkidis2023lexfiles}, and us-crimes~\cite{chalkidis2023lexfiles}. Details and statistics for these datasets are presented in Table~\ref{datasets}.

\subsection*{Results of In-Distributed Texts}
We create three subsets from the wiki dataset of varying sizes (i.e., wiki-small, wiki-large, and wiki-xl) and use these subsets to fine-tune the attack models, resulting in three distinct versions of the attack model. The performance of these attack models is then assessed using held-out texts from the wiki dataset. The experimental results presented in Table~\ref{main} illustrate that \emph{the size of the training datasets and the models have a considerable influence on the reconstruction accuracy}. To elaborate, 
regardless of the attack model employed, \emph{it's found that larger embedding models, such as BGE-Large-en and E5-Large-v2, enable more effective text reconstruction compared to others like SimCSE}. This is attributed to the strong expressivity of the large target embedding model, which allows it to retain more semantic information from the original text, proving beneficial for the embedding's application in subsequent tasks. Moreover, \emph{provided that the attack model is adequately fine-tuned and the embedding model is expressive enough, the accuracy of text reconstruction improves as the size of the attack model increases}. This improvement is reflected in the table's last two columns, showing higher accuracy as the attack model progresses from GPT2 to GPT2\_large, and finally to GPT2\_xl, attributed to the improved generative capabilities of larger models. Additionally, \emph{adequately fine-tuning the large language models is a prerequisite for their effectiveness in text reconstruction tasks. When the embeddings are less informative, fine-tuning the attack model demands a larger amount of training data}. This is highlighted by the lesser performance of GPT2\_xl compared to GPT2 when fine-tuning with wiki-small, which reverses after fine-tuning with Wiki-xl. Moreover, GPT2\_xl outperforms GPT2 in reconstructing text from SimCSE's embeddings as the fine-tuning dataset shifts from ``Wiki-large" to ``Wiki-xl".

To summarize, \textbf{a larger attack model, when fine-tuned with an increased amount of training data, is capable of more accurately reconstructing texts from the embeddings generated by target embedding models with higher expressivity}. Hence, a straightforward approach for safeguarding privacy involves not disclosing the original dataset when publishing its embedding database. Nonetheless, it remains to be investigated whether an attack model, fine-tuned on datasets with varying distributions, would be effective.

\begin{table*}[t]
\centering
\caption{Reconstruction attack performance on different datasets. The best results are highlighted in bold. $*$ represents that the advantage of the best-performed attack model over other models is statistically significant (\textit{p}-value $<$ 0.05).}
\scalebox{0.79}{
\begin{tabular}{ll|cc|cc|cc}
\hline
\multicolumn{2}{l|}{Test dataset}           & \multicolumn{2}{c|}{wiki\_bio} & \multicolumn{2}{c|}{cc\_news} & \multicolumn{2}{c}{pile\_pubmed} \\
Target model            & Attack model   & BLEU-1                  & ROUGE-1                 & BLEU-1                  & ROUGE-1                 & BLEU-1                  & ROUGE-1                \\
\hline
\multirow{3}{*}{SimCSE}       & GPT2        & 0.5428±0.0025          & 0.5596±0.0022          & 0.3881±0.0012          & 0.4487±0.0011          & 0.3623±0.0015          & 0.3976±0.0014          \\
                              & GPT2\_large & 0.5859±0.0011          & 0.6272±0.0015          & 0.4314±0.0015          & 0.5020±0.0014          & 0.4054±0.0009          & 0.4427±0.0011          \\
                              & GPT2\_xl    & \textbf{0.5878±0.0012}$^*$ & \textbf{0.6329±0.0016}$^*$ & \textbf{0.4355±0.0010}$^*$ & \textbf{0.5084±0.0008}$^*$ & \textbf{0.4133±0.0013}$^*$ & \textbf{0.4505±0.0010}$^*$ \\
\hline
\multirow{3}{*}{BGE-Large-en} & GPT2        & 0.4773±0.0025          & 0.5327±0.0020          & 0.3906±0.0014          & 0.4314±0.0013          & 0.3297±0.0013          & 0.3581±0.0011          \\
                              & GPT2\_large & 0.5497±0.0018          & 0.6015±0.0009          & 0.4339±0.0009          & 0.4867±0.0012          & 0.3819±0.0013          & 0.4074±0.0014          \\
                              & GPT2\_xl    & \textbf{0.5652±0.0030}$^*$ & \textbf{0.6200±0.0015}$^*$ & \textbf{0.4480±0.0008}$^*$ & \textbf{0.5038±0.0006}$^*$ & \textbf{0.3955±0.0019}$^*$ & \textbf{0.4218±0.0017}$^*$ \\
\hline
\multirow{3}{*}{E5-Large-v2}  & GPT2        & 0.5312±0.0015          & 0.5532±0.0014          & 0.4065±0.0009          & 0.4428±0.0010          & 0.3673±0.0012          & 0.3995±0.0009          \\
                              & GPT2\_large & 0.5695±0.0014          & 0.6206±0.0017          & 0.4521±0.0013          & 0.5006±0.0013          & 0.4174±0.0006          & 0.4523±0.0007          \\
                              & GPT2\_xl    & \textbf{0.5823±0.0012}$^*$ & \textbf{0.6354±0.0017}$^*$ & \textbf{0.4645±0.0014}$^*$ & \textbf{0.5173±0.0015}$^*$ & \textbf{0.4316±0.0009}$^*$ & \textbf{0.4683±0.0009}$^*$ \\
\bottomrule
\multicolumn{2}{l|}{Test dataset}           & \multicolumn{2}{c|}{triage} & \multicolumn{2}{c|}{us\_crimes} & \multicolumn{2}{c}{cjeu\_terms} \\
Target model            & Attack model   & BLEU-1                  & ROUGE-1                 & BLEU-1                  & ROUGE-1                 & BLEU-1                  & ROUGE-1                \\
\hline
\multirow{3}{*}{SimCSE}       & GPT2        & 0.0932±0.0007          & 0.1555±0.0010          & 0.3092±0.0006          & 0.3238±0.0008          & 0.3485±0.0010          & 0.3646±0.0009          \\
                              & GPT2\_large & 0.1188±0.0006          & 0.1756±0.0006          & \textbf{0.3268±0.0006}$^*$ & 0.3406±0.0003          & 0.3755±0.0012          & \textbf{0.3954±0.0008}$^*$ \\
                              & GPT2\_xl    & \textbf{0.1299±0.0006}$^*$ & \textbf{0.2004±0.0010}$^*$ & 0.3226±0.0005          & \textbf{0.3429±0.0006}$^*$ & \textbf{0.3739±0.0007}$^*$ & 0.3914±0.0006          \\
\hline
\multirow{3}{*}{BGE-Large-en} & GPT2        & 0.0828±0.0007          & 0.1072±0.0008          & 0.2705±0.0008          & 0.2834±0.0010          & 0.3271±0.0008          & 0.3378±0.0009          \\
                              & GPT2\_large & \textbf{0.1376±0.0008}$^*$ & \textbf{0.1659±0.0009}$^*$ & 0.2755±0.0008          & 0.3155±0.0004          & \textbf{0.3639±0.0016}$^*$ & \textbf{0.3785±0.0014}$^*$ \\
                              & GPT2\_xl    & 0.1232±0.0008          & 0.1640±0.0007          & \textbf{0.2775±0.0008}$^*$ & \textbf{0.3207±0.0006}$^*$ & 0.3522±0.0012          & 0.3794±0.0012          \\
\hline
\multirow{3}{*}{E5-Large-v2}  & GPT2        & 0.1451±0.0007          & 0.2265±0.0008          & 0.2825±0.0009          & 0.2938±0.0010          & 0.3418±0.0018          & 0.3668±0.0016          \\
                              & GPT2\_large & \textbf{0.2272±0.0004}$^*$ & \textbf{0.3172±0.0007}$^*$ & 0.3066±0.0007          & 0.3308±0.0006          & 0.3621±0.0013          & 0.3980±0.0012          \\
                              & GPT2\_xl    & 0.2230±0.0008          & 0.3177±0.0007          & \textbf{0.3164±0.0007}$^*$ & \textbf{0.3459±0.0005}$^*$ & \textbf{0.3688±0.0009}$^*$ & \textbf{0.4101±0.0011}$^*$ \\
\hline
\end{tabular}
}
\label{unware}
\end{table*}

\subsection*{Results of Out-of-Distributed Texts}
To address the unresolved question, we assume that the attacker model is trained on the Wiki dataset with more rigorous descriptions of world knowledge, yet the texts used for testing do not originate from this dataset. This implies that the distribution of the texts used for testing differs from that of the texts used for training. We assess the reconstruction capability of this attack model on sample texts from six other datasets: wiki\_bio, cc\_news, pile\_pubmed, triage, us\_crimes, and cjeu\_terms. The results presented in Table~\ref{unware} show that \emph{the attack model retains the capability to accurately reconstruct texts from the embeddings, even with texts derived from different distributions than its training data}. In greater detail, the best reconstruction accuracy of the attack model on texts from 5 out of 6 datasets is equal to or even exceeds that of a model fine-tuned on wiki-small, a relatively small subset of Wikipedia. As a result, if we release embeddings for the wiki\_bio, cc\_news, pile\_pubmed, us\_crimes, and cjeu\_terms datasets, an attack model fine-tuned on the Wiki-xl dataset can extract semantic information from them with relatively high confidence. This suggests that simply withholding the original raw data does not prevent an attacker from reconstructing the original text information from their released embeddings.

\begin{figure}[htbp]
\centering
\begin{minipage}[t]{0.48\linewidth}
\centering
\includegraphics[width=3.5cm]{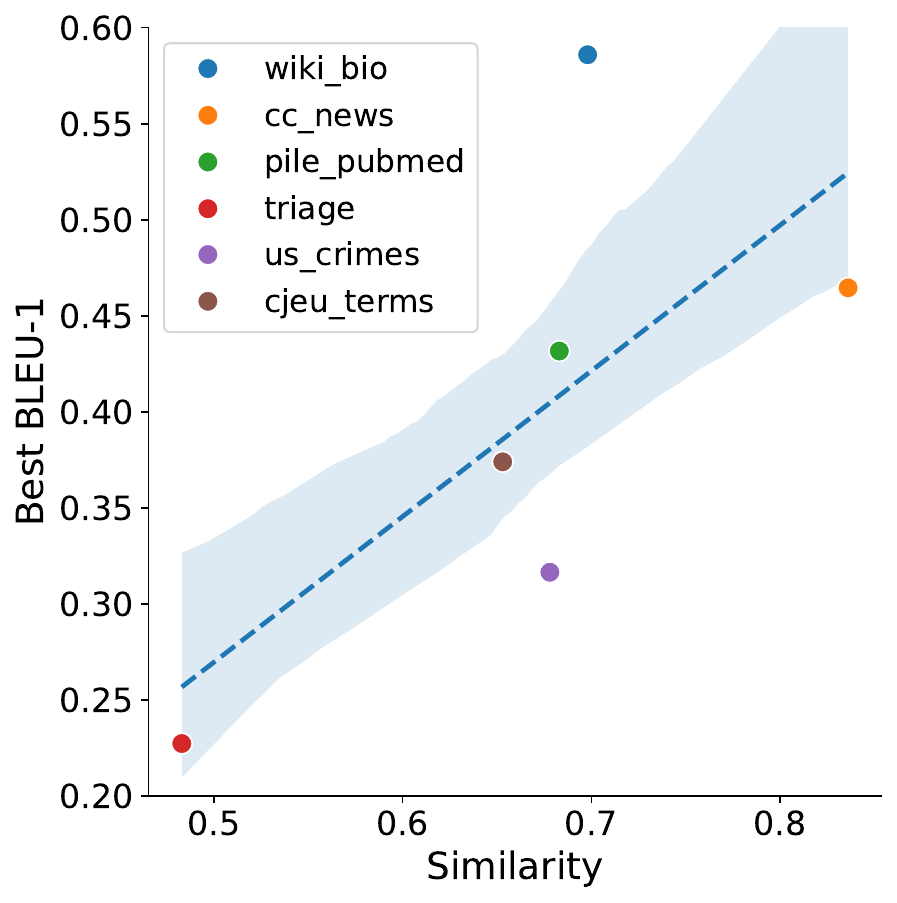}
\end{minipage}
\begin{minipage}[t]{0.48\linewidth}
\centering
\includegraphics[width=3.5cm]{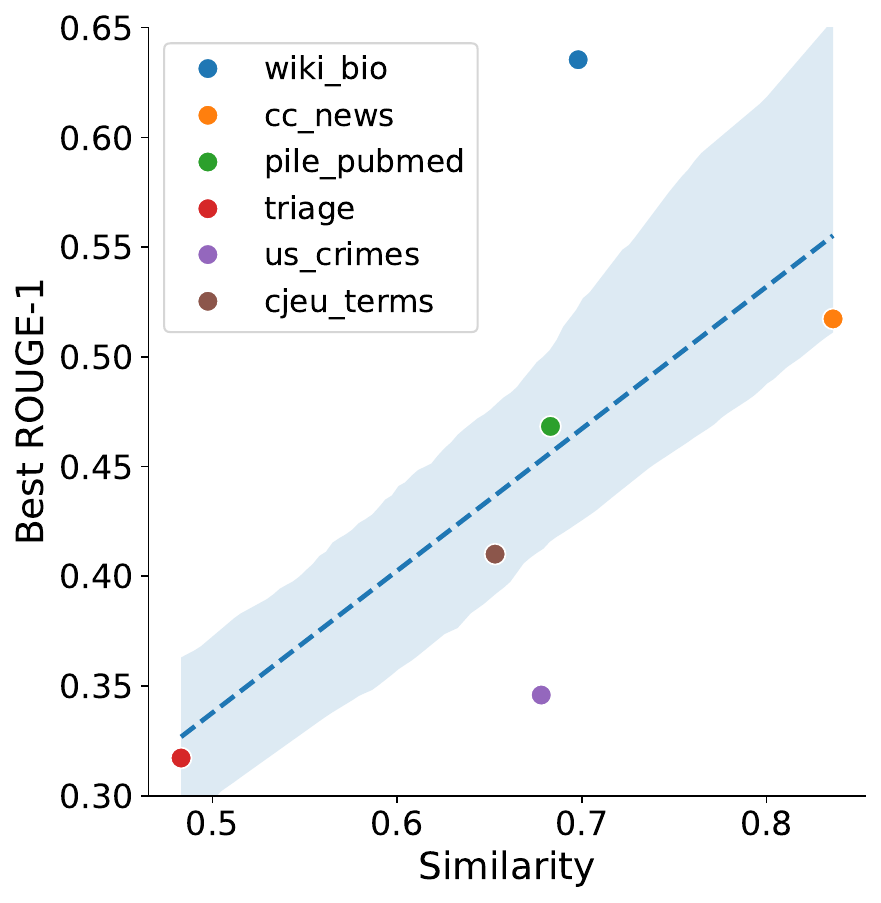}
\end{minipage}
\caption{The similarity between the evaluation datasets and the Wiki dataset v.s. the best reconstruction performance.}
\label{similarity}
\end{figure}

To understand which kind of text data is more easily recovered from text embedding based on the attack model fine-tuned with Wiki-xl, we also analyze the similarity between the six evaluation datasets and Wiki based on previous works~\cite{kilgarriff2001comparing}. The results reported in figure~\ref{similarity} show that \emph{texts from evaluation datasets with higher similarity to the training data are reconstructed more accurately}. To elaborate, Wiki-bio, which compiles biographies from Wikipedia, shares the same origin as the training dataset. Despite covering different content, the language style is very similar. Consequently, the quality of the attack's text reconstruction for this dataset is the highest. In other words, simply withholding the original text associated with embeddings does not adequately safeguard sensitive information from being extracted, as fine-tuning the attack model with datasets that are similar in terms of language style or content can elevate the risk of privacy breaches. 

\begin{figure}[htbp]
  \centering
  \includegraphics[width=\linewidth]{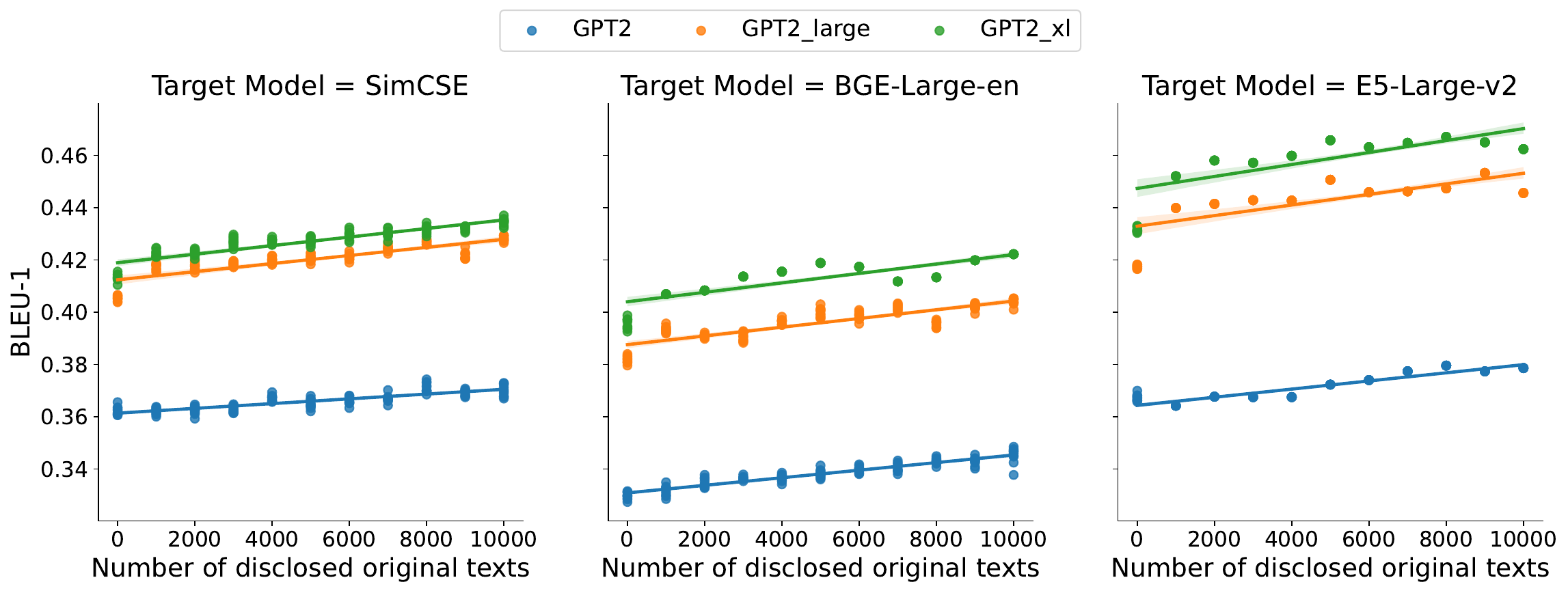}
  \quad
  \quad
  \includegraphics[width=\linewidth]{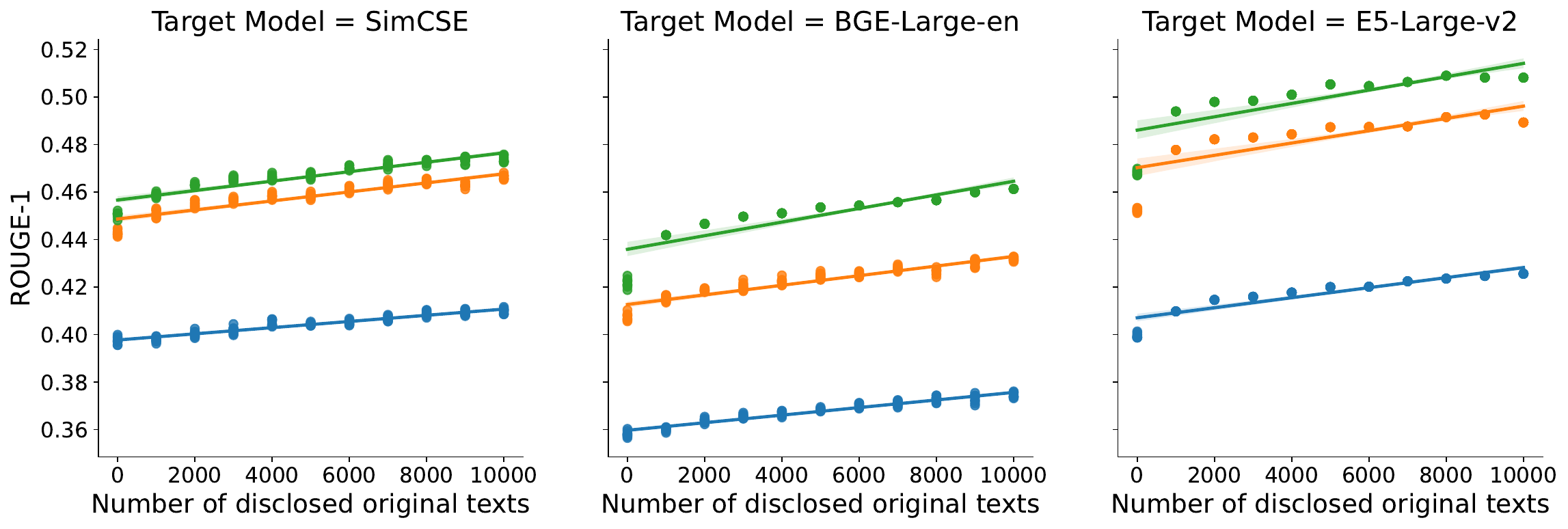}
  \vspace{-1em}
  \caption{\textbf{Impact of disclosed original texts volume on text reconstruction accuracy.} Each column represents a different target embedding model. The first and second rows represent the reconstruction performance concerning the BLEU-1 and ROUGE-1 metrics, respectively.}
  \label{few-shot}
\end{figure}

The disclosure of even a small collection of original texts significantly amplifies the risk, as illustrated in figure~\ref{few-shot} which presents results from the pile-pubmed dataset where the attack model undergoes further fine-tuning with these original texts. The availability of more original texts linked to the embedding directly correlates with an increased risk of sensitive information leakage. Specifically, the BLEU-1 score for GPT2-xl against the BGE-large-en embedding model sees a 2\% increase when the attack model is supplemented with 10k original texts. Notably, even with the use of a limited amount of target data (1K samples), the improvements in BLEU-1 score are considerable.

In summary, \textbf{concealing the datasets of published embeddings does not effectively prevent information leakage from these embeddings. This is because their underlying information can still be extracted by an attack model that has been fine-tuned on datasets of similar style or content. Furthermore, revealing even a small number of samples can significantly improve the extraction accuracy.}

\subsection*{Results of Varying Text Lengths}
Given GPT's capability to generate outputs of varying lengths with consistent meanings, exploring how text length impacts reconstruction quality becomes pertinent. With the average text length in the Wiki dataset being 21.32, as noted in Table~\ref{datasets}, we selected three subsets: Wiki-base, Wiki-medium, and Wiki-long, each with 5,000 samples but average lengths of 20, 40, and 80 words, respectively. Among them, Wiki-medium and Wiki-long are created by extending texts in Wiki-base via the GPT4 API~\cite{achiam2023gpt} accessible by OpenAI. 
\begin{figure}[h]
  \centering
  \includegraphics[width=\linewidth]{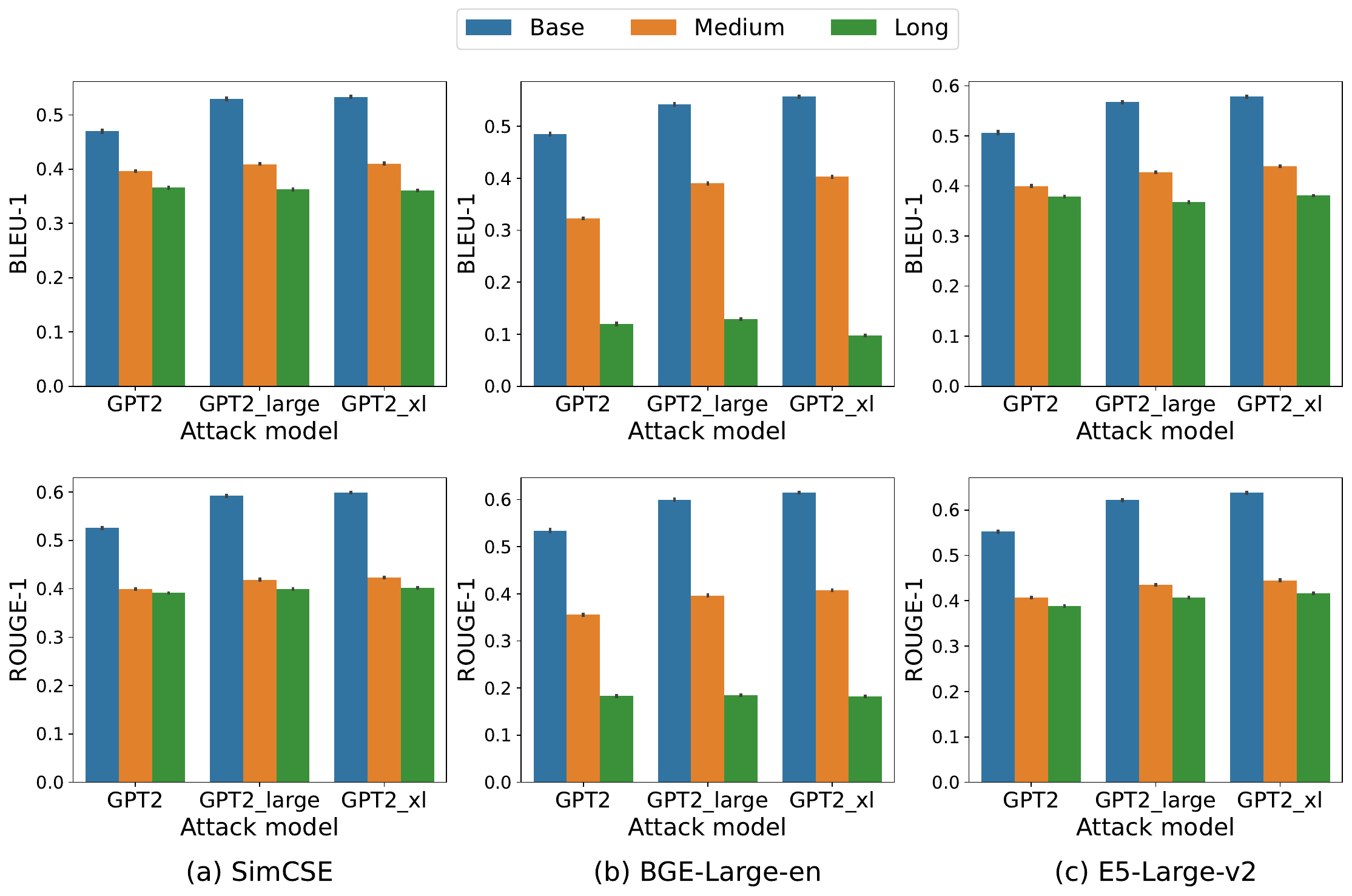}
  \caption{\textbf{Influence of the text length.} Error bars represent the mean reconstruction accuracy with 95\% confidence intervals obtained from 10 independent trials, and each column corresponds to a different target embedding model.}
  \label{text-len}
\end{figure}

The results of text reconstruction are depicted in figure~\ref{text-len}. 
It's evident that embeddings from shorter texts are more susceptible to being decoded, posing a greater risk to privacy. For example, the ROUGE-1 score of GPT2-xl on BGE-Large-en fell by over 43.3\% as the test text length increased from Wiki-base to Wiki-long. This decline can be attributed to the fixed length of text embeddings, which remain constant regardless of the original text's length. Consequently, embeddings of longer texts, which encapsulate more information within the same embedding size, make accurate text recovery more formidable. Therefore, \textbf{extending the original texts could potentially fortify the security of released embeddings}.

\begin{table*}[]
\centering
\caption{Attribute inference attack performance (accuracy) when using bge-large-en as the embedding model. The best results are highlighted in bold. $*$ denotes that the advantage of the best-performed attack model over other models is statistically significant (\textit{p}-value $<$ 0.05). 
The last row provides experimental results where the attacker has unrestricted access to the target embedding model and compares the embeddings of original text and candidate attributes. Other results are based on the attacker's usage of an external embedding model to calculate the similarities between reconstructed text and candidate attributes.}
\scalebox{0.78}{
\begin{tabular}{ll|ccc|cc|c|c}
\hline
\multicolumn{2}{l|}{Dataset}             & \multicolumn{3}{c|}{wiki-bio}                                                                       & \multicolumn{2}{c|}{triage}                                            & cjeu\_terms                     & us\_crimes                           \\
Similarity Model                     & Attack Model                     & nationality                     & birth\_date                     & occupation                     & disposition                     & blood\_pressure                     & legal\_term                     & criminal\_charge                     \\
\hline
\multirow{3}{*}{SimCSE}       & GPT2        & 0.875±0.005          & 0.546±0.008          & 0.869±0.005          & 0.506±0.001          & 0.135±0.004          & 0.395±0.010          & 0.377±0.003          \\
                              & GPT2\_large & 0.882±0.003          & 0.579±0.011          & \textbf{0.878±0.004} & 0.506±0.001          & 0.119±0.006          & 0.405±0.006          & 0.396±0.004          \\
                              & GPT2\_xl    & \textbf{0.886±0.003}$^*$ & \textbf{0.596±0.009}$^*$ & \textbf{0.878±0.005} & \textbf{0.514±0.002}$^*$ & \textbf{0.137±0.003} & \textbf{0.428±0.008}$^*$ & \textbf{0.407±0.004}$^*$ \\
\hline
\multirow{3}{*}{BGE-Large-en} & GPT2        & 0.927±0.005          & 0.525±0.009          & 0.913±0.006          & \textbf{0.505±0.002} & 0.195±0.004          & 0.496±0.008          & 0.470±0.005          \\
                              & GPT2\_large & 0.937±0.005          & 0.560±0.009          & 0.919±0.003          & 0.504±0.001          & 0.238±0.005          & 0.538±0.004          & 0.510±0.005          \\
                              & GPT2\_xl    & \textbf{0.941±0.003}$^*$ & \textbf{0.581±0.008}$^*$ & \textbf{0.922±0.005} & 0.504±0.001          & \textbf{0.254±0.007}$^*$ & \textbf{0.551±0.006}$^*$ & \textbf{0.527±0.003}$^*$ \\
\hline
\multirow{3}{*}{E5-Large-v2}  & GPT2        & 0.932±0.004          & 0.670±0.008          & 0.927±0.005          & 0.514±0.002          & 0.206±0.005          & 0.492±0.010          & 0.465±0.004          \\
                              & GPT2\_large & 0.940±0.003          & 0.729±0.008          & 0.938±0.003          & \textbf{0.521±0.001} & 0.229±0.003          & 0.544±0.007          & 0.506±0.006          \\
                              & GPT2\_xl    & \textbf{0.941±0.003} & \textbf{0.756±0.008}$^*$ & \textbf{0.940±0.005} & \textbf{0.521±0.002} & \textbf{0.230±0.004} & \textbf{0.545±0.006} & \textbf{0.524±0.006}$^*$ \\
\hline
BGE-Large-en                  & None        & 0.953          & 0.742          & 0.950          & 0.538          & 0.431          & 0.764          & 0.716          \\ 

\hline
\end{tabular}}
\label{attr}
\end{table*}

\subsection{Evaluation of Sensitive Attribute Prediction}

\begin{figure}[htbp]
    \centering
    \includegraphics[width=\columnwidth]{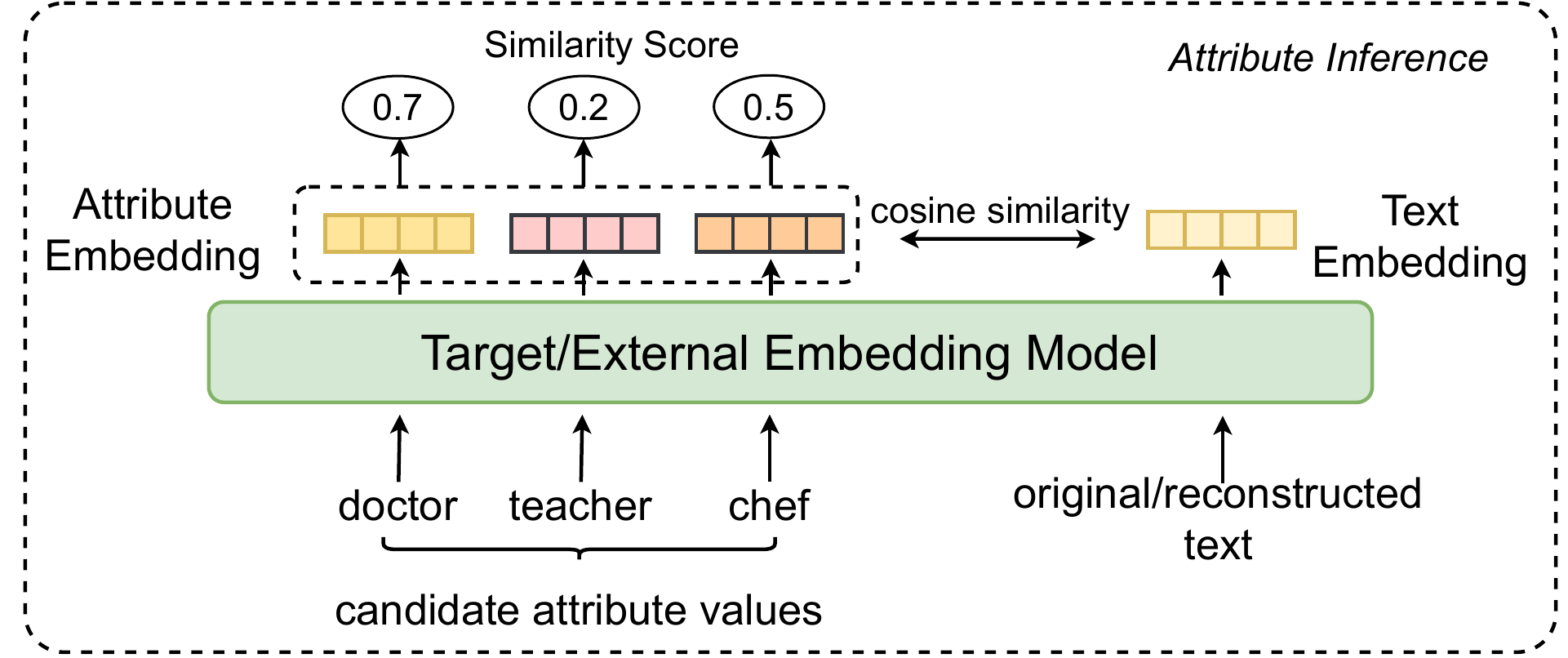}
    \vspace{-1em}
    \caption{\textbf{The inference framework of sensitive attributes.} The attacker employs the same embedding model to convert original text and candidate attributes into embeddings. The attacker then identifies the attribute that exhibits the highest cosine similarity between its embedding and text embedding as sensitive information of the original text.}
    \label{workflow2}
\end{figure}

\subsection*{Settings}
In contrast to text reconstruction, our primary concern is whether the attacker can extract specific sensitive information from the text embeddings. The task of predicting sensitive attributes more clearly illustrates the issue of privacy leakage through embeddings, compared to the task of reconstructing text. Initially, we pinpoint sensitive or crucial data within the datasets: wiki-bio, triage, cjeu-terms, and us-crimes. In particular, we examined patient dispositions and blood pressure readings in the triage dataset, and in the wiki-bio dataset, we focused on individuals' nationality, birthdate, and profession. Additionally, we looked into criminal charges in the us\_crimes dataset and considered legal terminology as important information in the cjeu\_terms dataset. Given the extensive variety of attributes and the scarcity of labeled data for each, it's impractical for the privacy attacker to train a dedicated model for each sensitive attribute, unlike the approach taken in previous studies~\cite{song2020information}. Therefore, we predicted the sensitive information from text embedding by selecting the attribute that exhibits the highest cosine similarity between text embedding and its embedding. This approach is effective across various attributes and does not necessitate training with supervised data. However, a challenge arises because the text describing the attribute is often very brief (sometimes just a single word), and the target embedding model may refuse to produce embeddings for such short texts due to concerns about embedding theft~\cite{feyisetan2021private, liu2019shared}. As a result, it becomes difficult to represent the original text and the attribute value within the same embedding space. To overcome this, we introduce an external embedding model to serve as an intermediary. This external model is responsible for embedding both the attribute and the reconstructed text, which has been derived from text embeddings by the attack model. Consequently, texts and attributes are embedded within the same space via reconstructing texts from the text embeddings, allowing for accurate similarity measurement. The overall process for inferring attributes is depicted in figure~\ref{workflow2}. The outcomes of the attribute inference attack using various methods are presented in Table~\ref{attr}. The last row of this table, which relies on the similarity between attributes and original texts, lacks randomness in its measurement due to the direct comparison method employed. In contrast, the preceding rows, which are based on reconstructed texts, introduce randomness into the similarity measurement. This variability stems from employing a non-zero temperature setting, allowing for multiple independent text generations to produce diverse outputs.

\subsection*{Results}

The findings presented in Table~\ref{attr} reveal that \emph{sensitive information can be accurately deduced from text embeddings without the necessity for any training data, highlighting a significant risk of privacy leakage through embeddings}. Specifically, attributes such as nationality and occupation can be inferred with high precision (an accuracy of 0.94) even when using texts that have been reconstructed. This level of accuracy is attributed to the embeddings' ability to capture the rich semantic details of texts coupled with the attack model's strong generative capabilities. Remarkably, the accuracy of predictions made using an external embedding model on reconstructed texts is on par with those made using the target embedding model on original texts in several cases. For an equitable comparison, the external embedding model was set to be identical to the target embedding model, although, in practice, the specifics of the target embedding model might not be known. The inference accuracy improves when employing a larger attack model for text reconstruction and a more expressive embedding model. This outcome aligns with observations from text reconstruction tasks. Although the accuracy of attribute prediction with reconstructed text falls short in some instances compared to using original texts, the ongoing advancement in large language models is rapidly closing this gap. Hence, the continuous evolution of these models is likely to escalate the risks associated with privacy breaches, emphasizing the critical need for increased awareness and caution in this domain.

\section{Discussions and Limitations}
This study delves into the implications of large language models on embedding privacy, focusing on text reconstruction and sensitive information prediction tasks. Our investigation shows that as the capabilities of both the sophisticated attack foundation model and the target embedding model increase, so does the risk of sensitive information leakage through knowledge embeddings. Furthermore, the risk intensifies when the attack model undergoes fine-tuning with data mirroring the distribution of texts linked to the released embeddings.

To protect the privacy of knowledge embeddings, we propose several strategies based on our experimental findings:

\begin{itemize}
\item \textbf{Cease the disclosure of original texts tied to released embeddings}: Preventing the attack model from being fine-tuned with similar datasets can be achieved by introducing imperceptible noise into the texts or embeddings. This aims to widen the gap between the dataset of original or reconstructed texts and other analogous datasets.
\item \textbf{Extend the length of short texts before embedding}: Enhancing short texts into longer versions while preserving their semantic integrity can be accomplished using GPT-4 or other large language models with similar generative capacities.
\item \textbf{Innovate new privacy-preserving embedding models}: Develop embedding models capable of producing high-quality text embeddings that are challenging to reverse-engineer into the original text. This entails training models to minimize the cloze task loss while maximizing the reconstruction loss.
\end{itemize}

However, our study is not without limitations. Firstly, due to substantial training expenses, we did not employ an attack model exceeding 10 billion parameters, though we anticipate similar outcomes with larger models. Secondly, while we have quantified the impact of various factors such as model size, text length, and training volume on embedding privacy, and outlined necessary guidelines for its protection, we have not formulated a concrete defense mechanism against potential embedding reconstruction attacks. Currently, effective safeguards primarily rely on perturbation techniques and encryption methods. Perturbation strategies, while protective, can compromise the embedding's utility in subsequent applications, necessitating a balance between security and performance. Encryption methods, though secure, often entail considerable computational demands. Future work will explore additional factors influencing embedding privacy breaches and seek methods for privacy-preserving embeddings without sacrificing their utility or incurring excessive computational costs.

\section{Methods}
\subsection{Preliminary}
Prior to delving into the attack strategies, we will commence with the introduction of the language models and evaluation datasets.
\subsection*{Language Model}
A language model is a technique capable of assessing the probability of a sequence of words forming a coherent sentence. Traditional language models, such as statistical language models~\cite{bellegarda2004statistical, rosenfeld2000two} and grammar rule language models~\cite{jurafsky1995using, sawaf2000use}, rely on heuristic methods to predict word sequences. While these conventional approaches may achieve high predictive accuracy for limited or straightforward sentences within small corpora, they often struggle to provide precise assessments for the majority of other word combinations. With increasing demands for the precision of language model predictions, numerous researchers have advocated for neural network-based language models~\cite{mikolov2011extensions, de2015survey} trained on extensive datasets. The performance of neural network-based language models steadily increases as model parameters are increased and sufficient training data is received. 
Upon reaching a certain threshold of parameter magnitude, the language model transcends previous paradigms to become a Large Language Model (LLM). The substantial parameter count within LLMs facilitates the acquisition of extensive implicit knowledge from corpora, thereby fostering the emergence of novel, powerful capabilities to handle more complex language tasks, such as arithmetic operations~\cite{muffo2023evaluating} and multi-step reasoning~\cite{paranjape2023art}. These capabilities have enabled large language models to comprehend and resolve issues like humans, leading to their rising popularity in many aspects of society. However, the significant inference capacity of large language models may be leveraged by attackers to reconstruct private information from text embeddings, escalating the risk of privacy leakage in embeddings.

\subsection*{Datasets for Evaluation}
\begin{table}[]
\centering
\caption{Statistics of datasets}
\begin{tabular}{l|c|c|c}
\hline
Dataset      & Domain & \#sentences & avg. sentence len \\
\hline
wiki         & General & 4,010,000 & 21.32   \\
wiki\_bio    & General & 1,480     & 22.19   \\  
cc\_news     & News    & 5,000     & 21.39   \\
pile\_pubmed & Medical & 5,000     & 21.93   \\
triage       & Medical & 4668      & 54.10   \\
cjeu\_terms  & Legal   & 2127      & 118.96  \\
us\_crimes   & Legal   & 4518      & 181.28  \\
\hline
\end{tabular}
\label{datasets}
\end{table}

In this paper, we assess the risk of embedding privacy leaks on seven datasets, including wiki~\cite{saez2020wikimedia}, wiki-bio~\cite{DBLP:journals/corr/LebretGA16}, cc-news~\cite{Hamborg2017}, pile-pubmed~\cite{gao2020pile}, triage~\cite{li2021topic}, cjeu-terms~\cite{chalkidis2023lexfiles}, and us-crimes~\cite{chalkidis2023lexfiles}. Wiki collects a large amount of text from the Wikipedia website. Since it has been vetted by the public, the text of the wiki is both trustworthy and high-quality.
Wiki-bio contains Wikipedia biographies that include the initial paragraph of the biography as well as the tabular infobox. CC-News (CommonCrawl News dataset) is a dataset containing news articles from international news websites. It contains 708241 English-language news articles published between January 2017 and December 2019. Pile-PubMed is a compilation of published medical literature from Pubmed, a free biomedical literature retrieval system developed by the National Center for Biotechnology Information (NCBI). It has housed over 4000 biomedical journals from more than 70 countries and regions since 1966. Triage records the triage notes, the demographic information, and the documented symptoms of the patients during the triage phase in the emergency center. Cjeu-term and us-crimes are two datasets from the legal domain. Cjeu-term collects some legal terminologies of the European Union, while us-crimes gathers transcripts of criminal cases in the United States. 

For wiki, cc\_news, and pile\_pubmed datasets, we randomly sample data from the original sets instead of using the entire dataset because the original datasets are enormous. 
To collect the sentence texts for reconstruction, we utilize en\_core\_web\_trf~\cite{encore}, an open-source tool, to segment the raw data into sentences. Then, we cleaned the data and filtered out sentences that were too long or too short. The statistical characteristics of the processed datasets are shown in Table~\ref{datasets}.

\subsection*{Ethics Statement}
For possible safety hazards, we abstained from conducting attacks on commercial embedding systems, instead employing open-source embedding models. Additionally, the datasets we utilized are all publicly accessible and anonymized, ensuring no user identities are involved. To reduce the possibility of privacy leakage, we opt to recover more general privacy attributes like occupation and nationality in the attribute prediction evaluation rather than attributes that could be connected to a specific individual, such as phone number and address. The intent of our research is to highlight the increased danger of privacy leakage posed by large language models in embeddings, suggest viable routes to safeguard embedding privacy through experimental analysis, and stimulate the community to develop more secure embedding models.

\subsection{Threat Model}
\subsection*{Attack Goal}
This paper primarily investigates the extraction of private data from text embeddings. Given that such private data often includes extremely sensitive information such as phone numbers, addresses, and occupations, attackers have ample motivation to carry out these attacks. For instance, they might engage in illegal activities such as telecommunications fraud or unauthorized selling of personal data for economic gain. These practices pose significant threats to individual privacy rights, potentially lead to economic losses, increase social instability, and undermine trust mechanisms.

\subsection*{Attack Knowledge}
To extract private information from text embeddings, attackers require some understanding of the models responsible for generating these embeddings. Intuitively, the more detailed the attacker's knowledge of the target embedding model, including its internal parameters, training specifics, etc., the more potent the attack's efficacy. However, in real-world scenarios, to safeguard their intellectual property and commercial interests, target models often keep their internal information confidential, providing access to users solely through a query interface $\mathcal{O}$. 
Based on the above considerations, this study assumes that attackers only possess query permissions to the target embedding model, which responds with the corresponding text embedding based on the text inputted by the attacker. This query process does not reveal any internal information about the model. Furthermore, with the increasing popularity of large language models, numerous companies are opting to release their anonymized datasets for academic research purposes. Therefore, this study also assumes that attackers have the capability to gather specific open-source data (e.g., Wikipedia in our evaluation) and leverage interactions with target models to acquire associated text embeddings. Clearly, such low-knowledge attack settings help simulate real attack scenarios and more accurately assess the risks of embedding privacy leaks.

\subsection{Attack Methodology}
\subsection*{Attack Model Construction} 
Attackers reconstructed original text from text embedding by training an attack model. However, they lack prior knowledge about which architecture should be used for the attack model. When using neural networks as the attack model, it is challenging to decide which neural network architecture should be employed. If the architecture of the attack model is not as expressive and complex as the embedding model, then it is difficult to ensure that it can extract private information.

Considering the exceptional performance of large language models across various domains, particularly in text comprehension~\cite{cheng2023adapting} and information extraction~\cite{dagdelen2024structured}, employing them as attack models could be an appropriate choice. Based on these considerations, this study utilizes GPT-2 models of varying sizes as attack models, training them to reconstruct the original text from the embeddings. 

\textbf{Training set generation.} We start by retrieving the text embeddings from the collected open-source data $D$ using the query interface $\mathcal{O}$ of the target embedding model. 
\begin{equation}
e_w = \mathcal{O}(w)
\end{equation}
Then, we construct the training dataset $D_{train}$ for the attack model based on these embeddings and corresponding texts.
\begin{equation}
D_{train} = \{(d_w, w)|w \in D, d_w = (e_w, <EOS>)\}
\end{equation}
where EOS(End-of-Sentence) token is a special token appended to the text embedding to signify the end of the embedding input.

The attack model performs the opposite operation compared to the embedding model: for each sample $(d_w, w)\in D_{train}$, the attack model strives to recover the original text $w$ from the embedding $e_w$.

\textbf{Attack model training.} 
For each embedding input $d_w$, the attack model is trained to output the first token of the original text.
Subsequently, when the attacker provides $d_w$ along with the first $i-1$ tokens of the original text, the attack model is tasked with predicting the $i-th$ token of the original text with utmost accuracy. Formally, for a sample $(d_w, w)$, the text reconstruction loss of the attack model is as follows:
\begin{equation}
    L_{\theta}(d_w,w) = - {\sum}_{i=1}^{l} {\log P(x_i|d_w,w_{<i}},\theta)
\end{equation}
where $w_{<i}=(x_1, ..., x_{i-1})$ represents the first $i-1$ tokens of the original text $w$. $l$ denotes the length of $w$ and $\theta$ represents the parameter of the attack model. 
Therefore, the training loss for the attack models is the sum of the text reconstruction loss across all samples in the training dataset:
\begin{equation}
    L_{\theta} = - {\sum}_{(d_w, w) \in D_{train}} L_{\theta}(d_w, w)
\end{equation}

To evaluate the effectiveness of the attack models, we employ two tasks: text reconstruction and attribute prediction.

\subsection*{Text Reconstruction Task} 
\textbf{Reconstructed text generation.} In the text reconstruction task, the attack models aim to generate reconstructed text that closely resembles the original text. 
When generating a reconstructed text of length $l$, the attacker aims for the generated text to have the highest likelihood among all candidate texts, which is formalized as follows:
\begin{equation}
w^{*} = \mathop{\arg\max}\limits_{w=\{x_1, ..., x_l\}} P(x_1, ..., x_l|d_w,\theta)
\end{equation}
However, on the one hand, the number of candidate tokens for $x_i$ in the reconstructed text $w$ is large, often exceeding $10,000$ in our experiments. On the other hand, the number of candidate texts exponentially increases as the text length $l$ grows. As a result, it becomes infeasible to iterate through all candidate texts of length $l$ and select the one with the highest likelihood as the output. 
A viable solution is greedy selection, which involves choosing the candidate token with the highest likelihood while progressively constructing the reconstructed text.
\begin{equation}
x_{i}^{*} = \mathop{\arg\max}\limits_{x_i} P(x_i|d_w,\theta,w_{\textless i}^{*})
\end{equation}
\begin{equation}
w^{*}=(x_1^{*}, ..., x_l^{*})
\end{equation}
However, this approach may easily lead the generation process into local optima. To enhance the quality of the reconstructed text and improve generation efficiency, we employ beam search in the text reconstruction task.
In beam search, if the number of beams is $k$, the algorithm maintains $k$ candidates with the highest generation probability at each step. Specifically, in the initial state, for a given input text embedding $e_w$, the attack model first records $k$ initial tokens with the highest generation likelihood (ignoring eos token) as candidate texts of length $1$ ($\mathcal{C}_{1}^{*}$).
\begin{equation}
\mathcal{C}_{1}^{*} = \mathop{\arg\max}\limits_{\mathcal{C}_{1} \subset \mathcal{X}, |\mathcal{C}_{1}|=k} {\sum}_{x \in \mathcal{C}_{1}} P(x|d_w,\theta)
\end{equation}
where $\mathcal{X}$ represents the set of all tokens in the attack model.

Subsequently, these $k$ initial tokens are combined with any token in $\mathcal{X}$ to create a text set of length $2$. The attack model then iterates through these texts of length $2$ and selects $k$ texts with the highest generation likelihood as candidate texts of length $2$ ($\mathcal{C}_{2}^{*}$). 
\begin{equation}
\mathcal{C}_{2}^{*} = \mathop{\arg\max}\limits_{|\mathcal{C}_{2}|=k} {\sum}_{(x_1,x_2) \in \mathcal{C}_{2}} P(x_1,x_2|d_w,\theta) 
\end{equation}
\begin{equation}
\mathcal{C}_{2} \subset \{(x_1,x_2)|x_1 \in \mathcal{C}_{1}^{*},x_2 \in \mathcal{X}\}
\end{equation}
This process continues, incrementing the text length until the model generates an EOS token signaling the end of the generation process.

\textbf{Evaluation metric.} We adopt BLEU-1 and ROUGE-1 to evaluate the reconstruction performance of the attack model, measuring how similar the reconstructed text $w'$ is to the original text $w$. The formulas for these two metrics are as follows:

\begin{small}
\begin{equation}
    \text{BLEU-1} = BP \cdot \frac{\sum_{x \in \text{set}(w')} \min(\text{count}(x, w), \text{count}(x, w'))}{\sum_{x \in \text{set}(w')} \text{count}(x, w')}
\end{equation}
\begin{equation}
    \text{ROUGE-1} = \frac{{\sum}_{x \in \text{set}(w)} \min(\text{count}(x, w), \text{count}(x, w'))}{{\sum}_{x \in \text{set}(w)} \text{count}(x, w)}
\end{equation}
\end{small}

where $\text{set}(w)$ and $\text{set}(w')$ are the sets of all tokens in $w$ and $w'$. $\text{Count}(x, w)$ and $\text{count}(x, w')$ are the number of times $x$ appears in $w$ and $w'$, respectively. The brevity penalty (BP) is used to prevent short sentences from getting an excessively high BLEU-1 score. BLEU-1 primarily assesses the similarity between the reconstructed text and the original text, whereas ROUGE-1 places greater emphasis on the completeness of the reconstruction results and whether the reconstructed text can encompass all the information present in the original text.

\textbf{Dataset similarity calculation.} We assess the similarity between the evaluation datasets and the wiki dataset based on a simple character n-gram comparison~\cite{kilgarriff2001comparing}. 
Specifically, we employed the 5000 commonly used 4-gram characters in English as the feature set $\mathcal{F}$ of the dataset. Each dataset is then represented as a 5000-dimensional feature vector.
\begin{equation}
    \overrightarrow{F_{D}} = [\text{count}(f_1,D),..., \text{count}(f_{5000},D)] 
\end{equation}
where $f_i \in \mathcal{F}$ is a 4-gram character, and $\text{count}(f_i,D)$ is the number of times $f_i$ appears in dataset $D$.
Finally, we calculate the Spearman correlation coefficient between the feature vectors of the two datasets to quantify their similarity.
\begin{equation}
    \text{Sim}(D_{1}, D_{2}) = \text{Spearman}(\overrightarrow{F_{1}}, \overrightarrow{F_{2}})
\end{equation}
where $\overrightarrow{F_{1}}$ and $\overrightarrow{F_{2}}$ are feature vectors of $D_{1}$ and $D_{2}$, respectively. The Spearman coefficient ranges from -1 to 1, where a higher value indicates a greater similarity between the two corresponding datasets.

\subsection*{Attribute Prediction Task}
In the attribute prediction task, this study focuses on the attacker's ability to extract private information from the original text. We chose several private attributes from four datasets and evaluated the attack model's ability to infer the precise values of these private attributes from the released text embedding. For example, in the wiki-bio dataset, occupation is chosen as a private attribute. The attacker attempts to ascertain that the original text contains the private message ``doctor" by using the embedding of the sentence ``David is a doctor." 

Instead of training the attack model to perform the attribute prediction task, this study utilizes the embedding similarity between the text and the attribute value to determine the suggested attribute value of the original text. Its rationality is that the text contains relevant information about sensitive attributes, so their embeddings should be similar. The ideal approach would be to determine based on the similarity between embeddings of the original text and sensitive attribute embeddings. However, this poses challenges: (1) The original text is unknown. (2) Privacy attributes are often short texts, and in most cases, consist of only one word; such frequent anomalous (short) inputs might be considered malicious attempts and rejected. Therefore, this study (1) uses reconstructed text instead of the original text, and (2) employs an open-source external embedding model as a proxy to obtain embeddings instead of using the target embedding model.  
It's worth noting that this study did not directly search for privacy attributes in the reconstructed text due to potential inaccuracies in reconstructing privacy attributes, such as missing tokens or reconstructing synonyms of the attributes.

Specifically, the attacker initially acquires embeddings of the reconstructed text and sensitive attribute from their proxy embedding model, subsequently computing the cosine similarity between them, and ultimately selecting the attribute with the highest similarity as the prediction result. Formally, the attacker infers the sensitive attribute $w_v$ as follows:
\begin{equation}
    w_v = \mathop{\arg\max}\limits_{v \in \mathbb{C}_{v}} \frac{e_{w'} \cdot e_v}{|e_{w'}| |e_v|}
\end{equation}
where $\mathbb{C}_{v}$ is the set of candidate attribute values, $w_v$ is the predicted attribute value of the original text $w$. $e_{w'}$ and $e_v$ are the embedding vectors of reconstructed text $w'$ and attribute value $v$ with the aid of the external embedding model, respectively. We employ accuracy as the metric to evaluate the performance of the attack model on the attribute prediction task.

\bibliography{ref}


\begin{thebibliography}{55}


\ifx \showCODEN    \undefined \def \showCODEN     #1{\unskip}     \fi
\ifx \showDOI      \undefined \def \showDOI       #1{#1}\fi
\ifx \showISBNx    \undefined \def \showISBNx     #1{\unskip}     \fi
\ifx \showISBNxiii \undefined \def \showISBNxiii  #1{\unskip}     \fi
\ifx \showISSN     \undefined \def \showISSN      #1{\unskip}     \fi
\ifx \showLCCN     \undefined \def \showLCCN      #1{\unskip}     \fi
\ifx \shownote     \undefined \def \shownote      #1{#1}          \fi
\ifx \showarticletitle \undefined \def \showarticletitle #1{#1}   \fi
\ifx \showURL      \undefined \def \showURL       {\relax}        \fi
\providecommand\bibfield[2]{#2}
\providecommand\bibinfo[2]{#2}
\providecommand\natexlab[1]{#1}
\providecommand\showeprint[2][]{arXiv:#2}

\bibitem[coh({[n.\,d.]})]%
        {cohere}
 \bibinfo{year}{[n.\,d.]}\natexlab{}.
\newblock \bibinfo{title}{The Cohere Platform}.
\newblock
\newblock
\newblock
\shownote{\url{https://docs.cohere.com/docs/}}.


\bibitem[enc({[n.\,d.]})]%
        {encore}
 \bibinfo{year}{[n.\,d.]}\natexlab{}.
\newblock \bibinfo{title}{English transformer pipeline}.
\newblock
\newblock
\newblock
\shownote{\url{https://huggingface.co/spacy/en_core_web_trf/}}.


\bibitem[Achiam et~al\mbox{.}(2023)]%
        {achiam2023gpt}
\bibfield{author}{\bibinfo{person}{Josh Achiam}, \bibinfo{person}{Steven Adler}, \bibinfo{person}{Sandhini Agarwal}, \bibinfo{person}{Lama Ahmad}, \bibinfo{person}{Ilge Akkaya}, \bibinfo{person}{Florencia~Leoni Aleman}, \bibinfo{person}{Diogo Almeida}, \bibinfo{person}{Janko Altenschmidt}, \bibinfo{person}{Sam Altman}, \bibinfo{person}{Shyamal Anadkat}, {et~al\mbox{.}}} \bibinfo{year}{2023}\natexlab{}.
\newblock \showarticletitle{Gpt-4 technical report}.
\newblock \bibinfo{journal}{\emph{arXiv preprint arXiv:2303.08774}} (\bibinfo{year}{2023}).
\newblock


\bibitem[Al-Abdallah and Al-Taani(2017)]%
        {al2017arabic}
\bibfield{author}{\bibinfo{person}{Raed~Z Al-Abdallah} {and} \bibinfo{person}{Ahmad~T Al-Taani}.} \bibinfo{year}{2017}\natexlab{}.
\newblock \showarticletitle{Arabic single-document text summarization using particle swarm optimization algorithm}.
\newblock \bibinfo{journal}{\emph{Procedia Computer Science}}  \bibinfo{volume}{117} (\bibinfo{year}{2017}), \bibinfo{pages}{30--37}.
\newblock


\bibitem[AlBadani et~al\mbox{.}(2022)]%
        {albadani2022novel}
\bibfield{author}{\bibinfo{person}{Barakat AlBadani}, \bibinfo{person}{Ronghua Shi}, {and} \bibinfo{person}{Jian Dong}.} \bibinfo{year}{2022}\natexlab{}.
\newblock \showarticletitle{A novel machine learning approach for sentiment analysis on Twitter incorporating the universal language model fine-tuning and SVM}.
\newblock \bibinfo{journal}{\emph{Applied System Innovation}} \bibinfo{volume}{5}, \bibinfo{number}{1} (\bibinfo{year}{2022}), \bibinfo{pages}{13}.
\newblock


\bibitem[Andriopoulos and Pouwelse(2023)]%
        {andriopoulos2023augmenting}
\bibfield{author}{\bibinfo{person}{Konstantinos Andriopoulos} {and} \bibinfo{person}{Johan Pouwelse}.} \bibinfo{year}{2023}\natexlab{}.
\newblock \showarticletitle{Augmenting LLMs with Knowledge: A survey on hallucination prevention}.
\newblock \bibinfo{journal}{\emph{arXiv preprint arXiv:2309.16459}} (\bibinfo{year}{2023}).
\newblock


\bibitem[Axelsson and Skantze(2023)]%
        {axelsson2023using}
\bibfield{author}{\bibinfo{person}{Agnes Axelsson} {and} \bibinfo{person}{Gabriel Skantze}.} \bibinfo{year}{2023}\natexlab{}.
\newblock \showarticletitle{Using large language models for zero-shot natural language generation from knowledge graphs}.
\newblock \bibinfo{journal}{\emph{arXiv preprint arXiv:2307.07312}} (\bibinfo{year}{2023}).
\newblock


\bibitem[Bellegarda(2004)]%
        {bellegarda2004statistical}
\bibfield{author}{\bibinfo{person}{Jerome~R Bellegarda}.} \bibinfo{year}{2004}\natexlab{}.
\newblock \showarticletitle{Statistical language model adaptation: review and perspectives}.
\newblock \bibinfo{journal}{\emph{Speech communication}} \bibinfo{volume}{42}, \bibinfo{number}{1} (\bibinfo{year}{2004}), \bibinfo{pages}{93--108}.
\newblock


\bibitem[Chalkidis et~al\mbox{.}(2023)]%
        {chalkidis2023lexfiles}
\bibfield{author}{\bibinfo{person}{Ilias Chalkidis}, \bibinfo{person}{Nicolas Garneau}, \bibinfo{person}{Catalina Goanta}, \bibinfo{person}{Daniel~Martin Katz}, {and} \bibinfo{person}{Anders S{\o}gaard}.} \bibinfo{year}{2023}\natexlab{}.
\newblock \showarticletitle{LeXFiles and LegalLAMA: Facilitating English Multinational Legal Language Model Development}.
\newblock \bibinfo{journal}{\emph{arXiv preprint arXiv:2305.07507}} (\bibinfo{year}{2023}).
\newblock


\bibitem[Chang et~al\mbox{.}(2023)]%
        {chang2023survey}
\bibfield{author}{\bibinfo{person}{Yupeng Chang}, \bibinfo{person}{Xu Wang}, \bibinfo{person}{Jindong Wang}, \bibinfo{person}{Yuan Wu}, \bibinfo{person}{Kaijie Zhu}, \bibinfo{person}{Hao Chen}, \bibinfo{person}{Linyi Yang}, \bibinfo{person}{Xiaoyuan Yi}, \bibinfo{person}{Cunxiang Wang}, \bibinfo{person}{Yidong Wang}, {et~al\mbox{.}}} \bibinfo{year}{2023}\natexlab{}.
\newblock \showarticletitle{A survey on evaluation of large language models}.
\newblock \bibinfo{journal}{\emph{arXiv preprint arXiv:2307.03109}} (\bibinfo{year}{2023}).
\newblock


\bibitem[Cheng et~al\mbox{.}(2023)]%
        {cheng2023adapting}
\bibfield{author}{\bibinfo{person}{Daixuan Cheng}, \bibinfo{person}{Shaohan Huang}, {and} \bibinfo{person}{Furu Wei}.} \bibinfo{year}{2023}\natexlab{}.
\newblock \showarticletitle{Adapting large language models via reading comprehension}.
\newblock \bibinfo{journal}{\emph{arXiv preprint arXiv:2309.09530}} (\bibinfo{year}{2023}).
\newblock


\bibitem[Coulter and Bensinger(2023)]%
        {coulter2023alphabet}
\bibfield{author}{\bibinfo{person}{Martin Coulter} {and} \bibinfo{person}{Greg Bensinger}.} \bibinfo{year}{2023}\natexlab{}.
\newblock \showarticletitle{Alphabet shares dive after Google AI chatbot Bard flubs answer in ad}.
\newblock \bibinfo{journal}{\emph{Reuters}} (\bibinfo{year}{2023}).
\newblock


\bibitem[Cui et~al\mbox{.}(2023)]%
        {cui2023chatlaw}
\bibfield{author}{\bibinfo{person}{Jiaxi Cui}, \bibinfo{person}{Zongjian Li}, \bibinfo{person}{Yang Yan}, \bibinfo{person}{Bohua Chen}, {and} \bibinfo{person}{Li Yuan}.} \bibinfo{year}{2023}\natexlab{}.
\newblock \showarticletitle{Chatlaw: Open-source legal large language model with integrated external knowledge bases}.
\newblock \bibinfo{journal}{\emph{arXiv preprint arXiv:2306.16092}} (\bibinfo{year}{2023}).
\newblock


\bibitem[Dagdelen et~al\mbox{.}(2024)]%
        {dagdelen2024structured}
\bibfield{author}{\bibinfo{person}{John Dagdelen}, \bibinfo{person}{Alexander Dunn}, \bibinfo{person}{Sanghoon Lee}, \bibinfo{person}{Nicholas Walker}, \bibinfo{person}{Andrew~S Rosen}, \bibinfo{person}{Gerbrand Ceder}, \bibinfo{person}{Kristin~A Persson}, {and} \bibinfo{person}{Anubhav Jain}.} \bibinfo{year}{2024}\natexlab{}.
\newblock \showarticletitle{Structured information extraction from scientific text with large language models}.
\newblock \bibinfo{journal}{\emph{Nature Communications}} \bibinfo{volume}{15}, \bibinfo{number}{1} (\bibinfo{year}{2024}), \bibinfo{pages}{1418}.
\newblock


\bibitem[De~Mulder et~al\mbox{.}(2015)]%
        {de2015survey}
\bibfield{author}{\bibinfo{person}{Wim De~Mulder}, \bibinfo{person}{Steven Bethard}, {and} \bibinfo{person}{Marie-Francine Moens}.} \bibinfo{year}{2015}\natexlab{}.
\newblock \showarticletitle{A survey on the application of recurrent neural networks to statistical language modeling}.
\newblock \bibinfo{journal}{\emph{Computer Speech \& Language}} \bibinfo{volume}{30}, \bibinfo{number}{1} (\bibinfo{year}{2015}), \bibinfo{pages}{61--98}.
\newblock


\bibitem[Duncan(1975)]%
        {duncan1975t}
\bibfield{author}{\bibinfo{person}{David~B Duncan}.} \bibinfo{year}{1975}\natexlab{}.
\newblock \showarticletitle{T tests and intervals for comparisons suggested by the data}.
\newblock \bibinfo{journal}{\emph{Biometrics}} (\bibinfo{year}{1975}), \bibinfo{pages}{339--359}.
\newblock


\bibitem[Feyisetan and Kasiviswanathan(2021)]%
        {feyisetan2021private}
\bibfield{author}{\bibinfo{person}{Oluwaseyi Feyisetan} {and} \bibinfo{person}{Shiva Kasiviswanathan}.} \bibinfo{year}{2021}\natexlab{}.
\newblock \showarticletitle{Private release of text embedding vectors}. In \bibinfo{booktitle}{\emph{Proceedings of the First Workshop on Trustworthy Natural Language Processing}}. \bibinfo{pages}{15--27}.
\newblock


\bibitem[Freitag and Al-Onaizan(2017)]%
        {freitag2017beam}
\bibfield{author}{\bibinfo{person}{Markus Freitag} {and} \bibinfo{person}{Yaser Al-Onaizan}.} \bibinfo{year}{2017}\natexlab{}.
\newblock \showarticletitle{Beam search strategies for neural machine translation}.
\newblock \bibinfo{journal}{\emph{arXiv preprint arXiv:1702.01806}} (\bibinfo{year}{2017}).
\newblock


\bibitem[Friedman and Broeck(2020)]%
        {friedman2020symbolic}
\bibfield{author}{\bibinfo{person}{Tal Friedman} {and} \bibinfo{person}{Guy Broeck}.} \bibinfo{year}{2020}\natexlab{}.
\newblock \showarticletitle{Symbolic querying of vector spaces: Probabilistic databases meets relational embeddings}. In \bibinfo{booktitle}{\emph{Conference on Uncertainty in Artificial Intelligence}}. PMLR, \bibinfo{pages}{1268--1277}.
\newblock


\bibitem[Gao et~al\mbox{.}(2020)]%
        {gao2020pile}
\bibfield{author}{\bibinfo{person}{Leo Gao}, \bibinfo{person}{Stella Biderman}, \bibinfo{person}{Sid Black}, \bibinfo{person}{Laurence Golding}, \bibinfo{person}{Travis Hoppe}, \bibinfo{person}{Charles Foster}, \bibinfo{person}{Jason Phang}, \bibinfo{person}{Horace He}, \bibinfo{person}{Anish Thite}, \bibinfo{person}{Noa Nabeshima}, {et~al\mbox{.}}} \bibinfo{year}{2020}\natexlab{}.
\newblock \showarticletitle{The {P}ile: An 800{GB} dataset of diverse text for language modeling}.
\newblock \bibinfo{journal}{\emph{arXiv preprint arXiv:2101.00027}} (\bibinfo{year}{2020}).
\newblock


\bibitem[Gao et~al\mbox{.}(2021)]%
        {gao2021simcse}
\bibfield{author}{\bibinfo{person}{Tianyu Gao}, \bibinfo{person}{Xingcheng Yao}, {and} \bibinfo{person}{Danqi Chen}.} \bibinfo{year}{2021}\natexlab{}.
\newblock \showarticletitle{{SimCSE}: Simple Contrastive Learning of Sentence Embeddings}. In \bibinfo{booktitle}{\emph{Empirical Methods in Natural Language Processing (EMNLP)}}.
\newblock


\bibitem[Goertzel(2014)]%
        {goertzel2014artificial}
\bibfield{author}{\bibinfo{person}{Ben Goertzel}.} \bibinfo{year}{2014}\natexlab{}.
\newblock \showarticletitle{Artificial general intelligence: concept, state of the art, and future prospects}.
\newblock \bibinfo{journal}{\emph{Journal of Artificial General Intelligence}} \bibinfo{volume}{5}, \bibinfo{number}{1} (\bibinfo{year}{2014}), \bibinfo{pages}{1}.
\newblock


\bibitem[Guo et~al\mbox{.}(2022)]%
        {guo2022manu}
\bibfield{author}{\bibinfo{person}{Rentong Guo}, \bibinfo{person}{Xiaofan Luan}, \bibinfo{person}{Long Xiang}, \bibinfo{person}{Xiao Yan}, \bibinfo{person}{Xiaomeng Yi}, \bibinfo{person}{Jigao Luo}, \bibinfo{person}{Qianya Cheng}, \bibinfo{person}{Weizhi Xu}, \bibinfo{person}{Jiarui Luo}, \bibinfo{person}{Frank Liu}, {et~al\mbox{.}}} \bibinfo{year}{2022}\natexlab{}.
\newblock \showarticletitle{Manu: a cloud native vector database management system}.
\newblock \bibinfo{journal}{\emph{arXiv preprint arXiv:2206.13843}} (\bibinfo{year}{2022}).
\newblock


\bibitem[Hamborg et~al\mbox{.}(2017)]%
        {Hamborg2017}
\bibfield{author}{\bibinfo{person}{Felix Hamborg}, \bibinfo{person}{Norman Meuschke}, \bibinfo{person}{Corinna Breitinger}, {and} \bibinfo{person}{Bela Gipp}.} \bibinfo{year}{2017}\natexlab{}.
\newblock \showarticletitle{news-please: A Generic News Crawler and Extractor}. In \bibinfo{booktitle}{\emph{Proceedings of the 15th International Symposium of Information Science}} (Berlin). \bibinfo{pages}{218--223}.
\newblock
\urldef\tempurl%
\url{https://doi.org/10.5281/zenodo.4120316}
\showDOI{\tempurl}


\bibitem[Hochreiter and Schmidhuber(1997)]%
        {hochreiter1997long}
\bibfield{author}{\bibinfo{person}{Sepp Hochreiter} {and} \bibinfo{person}{J{\"u}rgen Schmidhuber}.} \bibinfo{year}{1997}\natexlab{}.
\newblock \showarticletitle{Long short-term memory}.
\newblock \bibinfo{journal}{\emph{Neural computation}} \bibinfo{volume}{9}, \bibinfo{number}{8} (\bibinfo{year}{1997}), \bibinfo{pages}{1735--1780}.
\newblock


\bibitem[Jurafsky et~al\mbox{.}(1995)]%
        {jurafsky1995using}
\bibfield{author}{\bibinfo{person}{Daniel Jurafsky}, \bibinfo{person}{Chuck Wooters}, \bibinfo{person}{Jonathan Segal}, \bibinfo{person}{Andreas Stolcke}, \bibinfo{person}{Eric Fosler}, \bibinfo{person}{Gary Tajchaman}, {and} \bibinfo{person}{Nelson Morgan}.} \bibinfo{year}{1995}\natexlab{}.
\newblock \showarticletitle{Using a stochastic context-free grammar as a language model for speech recognition}. In \bibinfo{booktitle}{\emph{1995 International Conference on Acoustics, Speech, and Signal Processing}}, Vol.~\bibinfo{volume}{1}. IEEE, \bibinfo{pages}{189--192}.
\newblock


\bibitem[Kaddour et~al\mbox{.}(2023)]%
        {kaddour2023challenges}
\bibfield{author}{\bibinfo{person}{Jean Kaddour}, \bibinfo{person}{Joshua Harris}, \bibinfo{person}{Maximilian Mozes}, \bibinfo{person}{Herbie Bradley}, \bibinfo{person}{Roberta Raileanu}, {and} \bibinfo{person}{Robert McHardy}.} \bibinfo{year}{2023}\natexlab{}.
\newblock \showarticletitle{Challenges and applications of large language models}.
\newblock \bibinfo{journal}{\emph{arXiv preprint arXiv:2307.10169}} (\bibinfo{year}{2023}).
\newblock


\bibitem[Kilgarriff(2001)]%
        {kilgarriff2001comparing}
\bibfield{author}{\bibinfo{person}{Adam Kilgarriff}.} \bibinfo{year}{2001}\natexlab{}.
\newblock \showarticletitle{Comparing corpora}.
\newblock \bibinfo{journal}{\emph{International journal of corpus linguistics}} \bibinfo{volume}{6}, \bibinfo{number}{1} (\bibinfo{year}{2001}), \bibinfo{pages}{97--133}.
\newblock


\bibitem[Lebret et~al\mbox{.}(2016)]%
        {DBLP:journals/corr/LebretGA16}
\bibfield{author}{\bibinfo{person}{R{\'{e}}mi Lebret}, \bibinfo{person}{David Grangier}, {and} \bibinfo{person}{Michael Auli}.} \bibinfo{year}{2016}\natexlab{}.
\newblock \showarticletitle{Generating Text from Structured Data with Application to the Biography Domain}.
\newblock \bibinfo{journal}{\emph{CoRR}}  \bibinfo{volume}{abs/1603.07771} (\bibinfo{year}{2016}).
\newblock
\showeprint[arxiv]{1603.07771}
\urldef\tempurl%
\url{http://arxiv.org/abs/1603.07771}
\showURL{%
\tempurl}


\bibitem[Li et~al\mbox{.}(2023b)]%
        {li2023sentence}
\bibfield{author}{\bibinfo{person}{Haoran Li}, \bibinfo{person}{Mingshi Xu}, {and} \bibinfo{person}{Yangqiu Song}.} \bibinfo{year}{2023}\natexlab{b}.
\newblock \showarticletitle{Sentence Embedding Leaks More Information than You Expect: Generative Embedding Inversion Attack to Recover the Whole Sentence}.
\newblock \bibinfo{journal}{\emph{arXiv preprint arXiv:2305.03010}} (\bibinfo{year}{2023}).
\newblock


\bibitem[Li et~al\mbox{.}(2023a)]%
        {li2023halueval}
\bibfield{author}{\bibinfo{person}{Junyi Li}, \bibinfo{person}{Xiaoxue Cheng}, \bibinfo{person}{Wayne~Xin Zhao}, \bibinfo{person}{Jian-Yun Nie}, {and} \bibinfo{person}{Ji-Rong Wen}.} \bibinfo{year}{2023}\natexlab{a}.
\newblock \showarticletitle{Halueval: A large-scale hallucination evaluation benchmark for large language models}. In \bibinfo{booktitle}{\emph{Proceedings of the 2023 Conference on Empirical Methods in Natural Language Processing}}. \bibinfo{pages}{6449--6464}.
\newblock


\bibitem[Li et~al\mbox{.}(2021)]%
        {li2021topic}
\bibfield{author}{\bibinfo{person}{Yutong Li}, \bibinfo{person}{Ruoqing Zhu}, \bibinfo{person}{Annie Qu}, \bibinfo{person}{Han Ye}, {and} \bibinfo{person}{Zhankun Sun}.} \bibinfo{year}{2021}\natexlab{}.
\newblock \showarticletitle{Topic modeling on triage notes with semiorthogonal nonnegative matrix factorization}.
\newblock \bibinfo{journal}{\emph{J. Amer. Statist. Assoc.}} \bibinfo{volume}{116}, \bibinfo{number}{536} (\bibinfo{year}{2021}), \bibinfo{pages}{1609--1624}.
\newblock


\bibitem[Lin(2004)]%
        {lin2004rouge}
\bibfield{author}{\bibinfo{person}{Chin-Yew Lin}.} \bibinfo{year}{2004}\natexlab{}.
\newblock \showarticletitle{Rouge: A package for automatic evaluation of summaries}. In \bibinfo{booktitle}{\emph{Text summarization branches out}}. \bibinfo{pages}{74--81}.
\newblock


\bibitem[Liu et~al\mbox{.}(2019)]%
        {liu2019shared}
\bibfield{author}{\bibinfo{person}{Xuebo Liu}, \bibinfo{person}{Derek~F Wong}, \bibinfo{person}{Yang Liu}, \bibinfo{person}{Lidia~S Chao}, \bibinfo{person}{Tong Xiao}, {and} \bibinfo{person}{Jingbo Zhu}.} \bibinfo{year}{2019}\natexlab{}.
\newblock \showarticletitle{Shared-private bilingual word embeddings for neural machine translation}.
\newblock \bibinfo{journal}{\emph{arXiv preprint arXiv:1906.03100}} (\bibinfo{year}{2019}).
\newblock


\bibitem[Lu et~al\mbox{.}(2022)]%
        {lu2022learn}
\bibfield{author}{\bibinfo{person}{Pan Lu}, \bibinfo{person}{Swaroop Mishra}, \bibinfo{person}{Tanglin Xia}, \bibinfo{person}{Liang Qiu}, \bibinfo{person}{Kai-Wei Chang}, \bibinfo{person}{Song-Chun Zhu}, \bibinfo{person}{Oyvind Tafjord}, \bibinfo{person}{Peter Clark}, {and} \bibinfo{person}{Ashwin Kalyan}.} \bibinfo{year}{2022}\natexlab{}.
\newblock \showarticletitle{Learn to explain: Multimodal reasoning via thought chains for science question answering}.
\newblock \bibinfo{journal}{\emph{Advances in Neural Information Processing Systems}}  \bibinfo{volume}{35} (\bibinfo{year}{2022}), \bibinfo{pages}{2507--2521}.
\newblock


\bibitem[Mikolov et~al\mbox{.}(2011)]%
        {mikolov2011extensions}
\bibfield{author}{\bibinfo{person}{Tom{\'a}{\v{s}} Mikolov}, \bibinfo{person}{Stefan Kombrink}, \bibinfo{person}{Luk{\'a}{\v{s}} Burget}, \bibinfo{person}{Jan {\v{C}}ernock{\`y}}, {and} \bibinfo{person}{Sanjeev Khudanpur}.} \bibinfo{year}{2011}\natexlab{}.
\newblock \showarticletitle{Extensions of recurrent neural network language model}. In \bibinfo{booktitle}{\emph{2011 IEEE international conference on acoustics, speech and signal processing (ICASSP)}}. IEEE, \bibinfo{pages}{5528--5531}.
\newblock


\bibitem[Min et~al\mbox{.}(2023)]%
        {min2023recent}
\bibfield{author}{\bibinfo{person}{Bonan Min}, \bibinfo{person}{Hayley Ross}, \bibinfo{person}{Elior Sulem}, \bibinfo{person}{Amir Pouran~Ben Veyseh}, \bibinfo{person}{Thien~Huu Nguyen}, \bibinfo{person}{Oscar Sainz}, \bibinfo{person}{Eneko Agirre}, \bibinfo{person}{Ilana Heintz}, {and} \bibinfo{person}{Dan Roth}.} \bibinfo{year}{2023}\natexlab{}.
\newblock \showarticletitle{Recent advances in natural language processing via large pre-trained language models: A survey}.
\newblock \bibinfo{journal}{\emph{Comput. Surveys}} \bibinfo{volume}{56}, \bibinfo{number}{2} (\bibinfo{year}{2023}), \bibinfo{pages}{1--40}.
\newblock


\bibitem[Morris et~al\mbox{.}(2023)]%
        {morris2023text}
\bibfield{author}{\bibinfo{person}{John~X Morris}, \bibinfo{person}{Volodymyr Kuleshov}, \bibinfo{person}{Vitaly Shmatikov}, {and} \bibinfo{person}{Alexander~M Rush}.} \bibinfo{year}{2023}\natexlab{}.
\newblock \showarticletitle{Text embeddings reveal (almost) as much as text}.
\newblock \bibinfo{journal}{\emph{arXiv preprint arXiv:2310.06816}} (\bibinfo{year}{2023}).
\newblock


\bibitem[Muffo et~al\mbox{.}(2023)]%
        {muffo2023evaluating}
\bibfield{author}{\bibinfo{person}{Matteo Muffo}, \bibinfo{person}{Aldo Cocco}, {and} \bibinfo{person}{Enrico Bertino}.} \bibinfo{year}{2023}\natexlab{}.
\newblock \showarticletitle{Evaluating transformer language models on arithmetic operations using number decomposition}.
\newblock \bibinfo{journal}{\emph{arXiv preprint arXiv:2304.10977}} (\bibinfo{year}{2023}).
\newblock


\bibitem[Neelakantan et~al\mbox{.}(2022)]%
        {neelakantan2022text}
\bibfield{author}{\bibinfo{person}{Arvind Neelakantan}, \bibinfo{person}{Tao Xu}, \bibinfo{person}{Raul Puri}, \bibinfo{person}{Alec Radford}, \bibinfo{person}{Jesse~Michael Han}, \bibinfo{person}{Jerry Tworek}, \bibinfo{person}{Qiming Yuan}, \bibinfo{person}{Nikolas Tezak}, \bibinfo{person}{Jong~Wook Kim}, \bibinfo{person}{Chris Hallacy}, {et~al\mbox{.}}} \bibinfo{year}{2022}\natexlab{}.
\newblock \showarticletitle{Text and code embeddings by contrastive pre-training}.
\newblock \bibinfo{journal}{\emph{arXiv preprint arXiv:2201.10005}} (\bibinfo{year}{2022}).
\newblock


\bibitem[Papineni et~al\mbox{.}(2002)]%
        {papineni2002bleu}
\bibfield{author}{\bibinfo{person}{Kishore Papineni}, \bibinfo{person}{Salim Roukos}, \bibinfo{person}{Todd Ward}, {and} \bibinfo{person}{Wei-Jing Zhu}.} \bibinfo{year}{2002}\natexlab{}.
\newblock \showarticletitle{Bleu: a method for automatic evaluation of machine translation}. In \bibinfo{booktitle}{\emph{Proceedings of the 40th annual meeting of the Association for Computational Linguistics}}. \bibinfo{pages}{311--318}.
\newblock


\bibitem[Paranjape et~al\mbox{.}(2023)]%
        {paranjape2023art}
\bibfield{author}{\bibinfo{person}{Bhargavi Paranjape}, \bibinfo{person}{Scott Lundberg}, \bibinfo{person}{Sameer Singh}, \bibinfo{person}{Hannaneh Hajishirzi}, \bibinfo{person}{Luke Zettlemoyer}, {and} \bibinfo{person}{Marco~Tulio Ribeiro}.} \bibinfo{year}{2023}\natexlab{}.
\newblock \showarticletitle{Art: Automatic multi-step reasoning and tool-use for large language models}.
\newblock \bibinfo{journal}{\emph{arXiv preprint arXiv:2303.09014}} (\bibinfo{year}{2023}).
\newblock


\bibitem[Patil et~al\mbox{.}(2023)]%
        {patil2023survey}
\bibfield{author}{\bibinfo{person}{Rajvardhan Patil}, \bibinfo{person}{Sorio Boit}, \bibinfo{person}{Venkat Gudivada}, {and} \bibinfo{person}{Jagadeesh Nandigam}.} \bibinfo{year}{2023}\natexlab{}.
\newblock \showarticletitle{A Survey of Text Representation and Embedding Techniques in NLP}.
\newblock \bibinfo{journal}{\emph{IEEE Access}} (\bibinfo{year}{2023}).
\newblock


\bibitem[Pei et~al\mbox{.}(2019)]%
        {pei2019towards}
\bibfield{author}{\bibinfo{person}{Jing Pei}, \bibinfo{person}{Lei Deng}, \bibinfo{person}{Sen Song}, \bibinfo{person}{Mingguo Zhao}, \bibinfo{person}{Youhui Zhang}, \bibinfo{person}{Shuang Wu}, \bibinfo{person}{Guanrui Wang}, \bibinfo{person}{Zhe Zou}, \bibinfo{person}{Zhenzhi Wu}, \bibinfo{person}{Wei He}, {et~al\mbox{.}}} \bibinfo{year}{2019}\natexlab{}.
\newblock \showarticletitle{Towards artificial general intelligence with hybrid Tianjic chip architecture}.
\newblock \bibinfo{journal}{\emph{Nature}} \bibinfo{volume}{572}, \bibinfo{number}{7767} (\bibinfo{year}{2019}), \bibinfo{pages}{106--111}.
\newblock


\bibitem[Radford et~al\mbox{.}(2019)]%
        {radford2019language}
\bibfield{author}{\bibinfo{person}{Alec Radford}, \bibinfo{person}{Jeffrey Wu}, \bibinfo{person}{Rewon Child}, \bibinfo{person}{David Luan}, \bibinfo{person}{Dario Amodei}, \bibinfo{person}{Ilya Sutskever}, {et~al\mbox{.}}} \bibinfo{year}{2019}\natexlab{}.
\newblock \showarticletitle{Language models are unsupervised multitask learners}.
\newblock \bibinfo{journal}{\emph{OpenAI blog}} \bibinfo{volume}{1}, \bibinfo{number}{8} (\bibinfo{year}{2019}), \bibinfo{pages}{9}.
\newblock


\bibitem[Rosenfeld(2000)]%
        {rosenfeld2000two}
\bibfield{author}{\bibinfo{person}{Ronald Rosenfeld}.} \bibinfo{year}{2000}\natexlab{}.
\newblock \showarticletitle{Two decades of statistical language modeling: Where do we go from here?}
\newblock \bibinfo{journal}{\emph{Proc. IEEE}} \bibinfo{volume}{88}, \bibinfo{number}{8} (\bibinfo{year}{2000}), \bibinfo{pages}{1270--1278}.
\newblock


\bibitem[Saez-Trumper and Redi(2020)]%
        {saez2020wikimedia}
\bibfield{author}{\bibinfo{person}{Diego Saez-Trumper} {and} \bibinfo{person}{Miriam Redi}.} \bibinfo{year}{2020}\natexlab{}.
\newblock \showarticletitle{Wikimedia Public (Research) Resources}. In \bibinfo{booktitle}{\emph{Companion Proceedings of the Web Conference 2020}}. \bibinfo{pages}{311--312}.
\newblock


\bibitem[Sawaf et~al\mbox{.}(2000)]%
        {sawaf2000use}
\bibfield{author}{\bibinfo{person}{Hassan Sawaf}, \bibinfo{person}{Kai Sch{\"u}tz}, {and} \bibinfo{person}{Hermann Ney}.} \bibinfo{year}{2000}\natexlab{}.
\newblock \showarticletitle{On the use of grammar based language models for statistical machine translation}. In \bibinfo{booktitle}{\emph{Proceedings of the Sixth International Workshop on Parsing Technologies}}. \bibinfo{pages}{231--241}.
\newblock


\bibitem[Selva~Birunda and Kanniga~Devi(2021)]%
        {selva2021review}
\bibfield{author}{\bibinfo{person}{S Selva~Birunda} {and} \bibinfo{person}{R Kanniga~Devi}.} \bibinfo{year}{2021}\natexlab{}.
\newblock \showarticletitle{A review on word embedding techniques for text classification}.
\newblock \bibinfo{journal}{\emph{Innovative Data Communication Technologies and Application: Proceedings of ICIDCA 2020}} (\bibinfo{year}{2021}), \bibinfo{pages}{267--281}.
\newblock


\bibitem[Song and Raghunathan(2020)]%
        {song2020information}
\bibfield{author}{\bibinfo{person}{Congzheng Song} {and} \bibinfo{person}{Ananth Raghunathan}.} \bibinfo{year}{2020}\natexlab{}.
\newblock \showarticletitle{Information leakage in embedding models}. In \bibinfo{booktitle}{\emph{Proceedings of the 2020 ACM SIGSAC conference on computer and communications security}}. \bibinfo{pages}{377--390}.
\newblock


\bibitem[Wang et~al\mbox{.}(2022)]%
        {wang2022text}
\bibfield{author}{\bibinfo{person}{Liang Wang}, \bibinfo{person}{Nan Yang}, \bibinfo{person}{Xiaolong Huang}, \bibinfo{person}{Binxing Jiao}, \bibinfo{person}{Linjun Yang}, \bibinfo{person}{Daxin Jiang}, \bibinfo{person}{Rangan Majumder}, {and} \bibinfo{person}{Furu Wei}.} \bibinfo{year}{2022}\natexlab{}.
\newblock \showarticletitle{Text Embeddings by Weakly-Supervised Contrastive Pre-training}.
\newblock \bibinfo{journal}{\emph{arXiv preprint arXiv:2212.03533}} (\bibinfo{year}{2022}).
\newblock


\bibitem[Wankhade et~al\mbox{.}(2022)]%
        {wankhade2022survey}
\bibfield{author}{\bibinfo{person}{Mayur Wankhade}, \bibinfo{person}{Annavarapu Chandra~Sekhara Rao}, {and} \bibinfo{person}{Chaitanya Kulkarni}.} \bibinfo{year}{2022}\natexlab{}.
\newblock \showarticletitle{A survey on sentiment analysis methods, applications, and challenges}.
\newblock \bibinfo{journal}{\emph{Artificial Intelligence Review}} \bibinfo{volume}{55}, \bibinfo{number}{7} (\bibinfo{year}{2022}), \bibinfo{pages}{5731--5780}.
\newblock


\bibitem[Xiao et~al\mbox{.}(2023)]%
        {bge_embedding}
\bibfield{author}{\bibinfo{person}{Shitao Xiao}, \bibinfo{person}{Zheng Liu}, \bibinfo{person}{Peitian Zhang}, {and} \bibinfo{person}{Niklas Muennighoff}.} \bibinfo{year}{2023}\natexlab{}.
\newblock \bibinfo{title}{C-Pack: Packaged Resources To Advance General Chinese Embedding}.
\newblock
\newblock
\showeprint[arxiv]{2309.07597}~[cs.CL]


\bibitem[Zhang et~al\mbox{.}(2023)]%
        {zhang2023siren}
\bibfield{author}{\bibinfo{person}{Yue Zhang}, \bibinfo{person}{Yafu Li}, \bibinfo{person}{Leyang Cui}, \bibinfo{person}{Deng Cai}, \bibinfo{person}{Lemao Liu}, \bibinfo{person}{Tingchen Fu}, \bibinfo{person}{Xinting Huang}, \bibinfo{person}{Enbo Zhao}, \bibinfo{person}{Yu Zhang}, \bibinfo{person}{Yulong Chen}, {et~al\mbox{.}}} \bibinfo{year}{2023}\natexlab{}.
\newblock \showarticletitle{Siren's Song in the AI Ocean: A Survey on Hallucination in Large Language Models}.
\newblock \bibinfo{journal}{\emph{arXiv preprint arXiv:2309.01219}} (\bibinfo{year}{2023}).
\newblock


\bibitem[Zhu et~al\mbox{.}(2021)]%
        {zhu2021retrieving}
\bibfield{author}{\bibinfo{person}{Fengbin Zhu}, \bibinfo{person}{Wenqiang Lei}, \bibinfo{person}{Chao Wang}, \bibinfo{person}{Jianming Zheng}, \bibinfo{person}{Soujanya Poria}, {and} \bibinfo{person}{Tat-Seng Chua}.} \bibinfo{year}{2021}\natexlab{}.
\newblock \showarticletitle{Retrieving and reading: A comprehensive survey on open-domain question answering}.
\newblock \bibinfo{journal}{\emph{arXiv preprint arXiv:2101.00774}} (\bibinfo{year}{2021}).
\newblock


\end{thebibliography}
\bibliographystyle{ACM-Reference-Format}

\end{document}